\documentclass{llncs}
\usepackage{algorithmic}
\usepackage{algorithm}
\usepackage{graphicx}
\usepackage{amsmath}
\usepackage{amssymb}
\usepackage{mathtools}

\usepackage{graphicx}
\begin{document}
\title{Improving Autoregressive NLP Tasks via Modular Linearized Attention}
\titlerunning{Modular Linearized Attention}
%
\author{Victor Agostinelli, Lizhong Chen}

\institute{Oregon State University}
%
%
%
\maketitle              
\begin{abstract}
Various natural language processing (NLP) tasks necessitate models that are efficient and small based on their ultimate application at the edge or other resource-constrained environment. While prior research has reduced the size of these models, increasing computational efficiency without considerable performance impacts remains difficult, especially for autoregressive tasks. This paper proposes \textit{modular linearized attention (MLA)}, which combines multiple efficient attention mechanisms, including cosFormer \cite{zhen2022cosformer}, to maximize inference quality while achieving notable speedups. We validate this approach on several autoregressive NLP tasks, including speech-to-text neural machine translation (S2T NMT), speech-to-text simultaneous translation (SimulST), and autoregressive text-to-spectrogram, noting efficiency gains on TTS and competitive performance for NMT and SimulST during training and inference.



\end{abstract}

\section{Introduction}
Transformers \cite{vaswani17} have provided researchers and industry leaders with new ways to take advantage of sequential data in natural language processing (NLP), computer vision, and a number of other fields \cite{vision_transf,music_transf,protein_transf}. The construction and deployment of efficient transformers that address the bottleneck of classical, quadratic attention mechanisms is a long-standing effort. Associated efficiency gains, however, are often coupled with significant inference quality costs. Prior transformer development has focused on mitigating those costs in several ways, such as carefully limiting available context for each token \cite{image_transf,adaptive_span2019,memformer}, making assumptions about the condition of the intermediate $QK^T$ matrix and applying pattern-based attention \cite{longformer2020,sparsetransformer,bigbird2020}, or attempting to model the input matrices in a lower dimensional space \cite{linformer2020}. Unfortunately, the aforementioned methods often are limited in their scope and can perform poorly when attempting to generalize to new tasks, in many cases due to environmental assumptions.

While truly linearized attention, with no prior environmental assumptions, is possible via several newer methods \cite{performer2020,kathrapalous20}, these implementations often struggle to approach the inference quality of quadratic, softmax-based attention. cosFormer \cite{zhen2022cosformer} is a recently proposed, state-of-the-art attention linearization technique that focuses on providing a tight approximation of softmax functionality for attention mechanisms. It emphasizes token locality and is proven to maintain long-range dependencies. In spite of its excellent performance on a range of tasks, there are significant limitations associated with cosFormer. These include a lack of validation on decoder cross-attention and suffering from a training to evaluation environment mismatch concerning sequence length availability for most autoregressive tasks. Moreover, that aforementioned token locality emphasis necessarily limits the widespread applicability of cosFormer.

Noting the issues facing current efficient transformers and problems concerning the applicability of individual efficient attention mechanisms, we propose \textit{modular linearized attention (MLA)}, a new design method that combines multiple attention paradigms for different attention blocks in a tight search space. This method avoids the pitfalls of a ``one-size-fits-all'' approach to efficient transformer construction. To enable \textit{MLA} for cosFormer, we propose an augmentation to its general formulation to allow for its application to decoder cross-attention. Additionally, as cosFormer struggles with most autoregressive tasks with dynamic sequence lengths, we propose several techniques related to predicting the target sequence length. The effectiveness of our approach is demonstrated across multiple tasks, with efficiency gains of up to a 7\% increase in decoder throughput during inference for autoregressive text-to-spectrogram (TTS), inference quality gains of 0.6 BLEU for English to German (en-de) speech-to-text neural machine translation (S2T NMT), and even a perplexity reduction of 0.44 during training for English to German speech-to-text simultaneous translation (SimulST).

\section{Background and Motivation}
\subsection{Fundamentals of Transformers and Attention}
The core architecture of Vaswani et. al's \cite{vaswani17} original transformer has been studied exhaustively, so we will only review it briefly and focus on the attention-based elements. At a high level, a classical transformer is composed of an encoder stack and a decoder stack. Both encoder and decoder layers contain a self-attention block in addition to a feed-forward block, with every decoder layer also containing a cross-attention block. Every block is wrapped by a residual connection and normalization. 

Every attention block takes in a query $Q$ in $\mathbb{R}^{N_1 \times d_k}$, a key $K$ in $\mathbb{R}^{N_2 \times d_k}$, and a value $V$ in $\mathbb{R}^{N_2 \times d_v}$. Classically, a softmax operator is applied to the product of the query and key to further distance tokens that are less relevant to one another and generate a probability distribution. These attention blocks are usually multi-headed for heads $H$ and the original embedding space is divided between each head before being reformed by concatenating the head outputs. Each attention head projects the input matrices into this new sub-space via weight matrices $W^q_h$ in $\mathbb{R}^{d_k \times d_{kh}}$, $W^k_h$ in $\mathbb{R}^{d_k \times d_{kh}}$, and $W^v_h$ in $\mathbb{R}^{d_v \times d_{vh}}$ for some head $h$ in $H$. A final output projection layer is characterized by $W_O$ in $\mathbb{R}^{Hd_v \times d_{model}}$ and is applied to the concatenation of head outputs. General formulations for multi-head attention calculations of a classical transformer are provided by Equations \ref{eq:softmax_line} and \ref{eq:mha_conc}. 

\begin{equation}
a_h = softmax(\frac{QW^q_hK^TW^k_h}{\sqrt{d_{kh}}})W^v_hV
\label{eq:softmax_line}
\end{equation}

\begin{equation}
A_{mha} = concat(a_1, a_2, \dots, a_H)W_O
\label{eq:mha_conc}
\end{equation}


\subsection{Early Efficient Transformers}


The quadratic bottleneck of Vaswani et. al's attention mechanism would prove problematic for workloads with long sequences, sometimes precluding the deployment of a transformer-based model. A number of efficient attention mechanisms are inspired by convolutional elements \cite{transformerxl,set_transf2018,liu2018memcomp,image_transf,adaptive_span2019,memformer}, often using sliding window mechanisms that limit context for attention mechanisms. Child et. al's SparseTransformer \cite{sparsetransformer} exploits sparsity-based opportunities to reduce the computational complexity of attention mechanisms and is foundational for several similar schemes \cite{etc_ainslie2020,longformer2020,bigbird2020}. Alternatively, the viability of low-rank representations of the intermediate $QK^T$ matrix is underscored by Linformer \cite{linformer2020}. Another unique scheme is Reformer \cite{reformer2020}, which provides a  generalizable run-time complexity reduction via locality sensitive hashing (LSH) and achieves competitive results for a number of tasks. 

\vspace{-0.5em}

\subsection{Linear Transformers}
\begin{figure}[h]
    \vspace{-2em}
    \centering
    \includegraphics[scale=0.3]{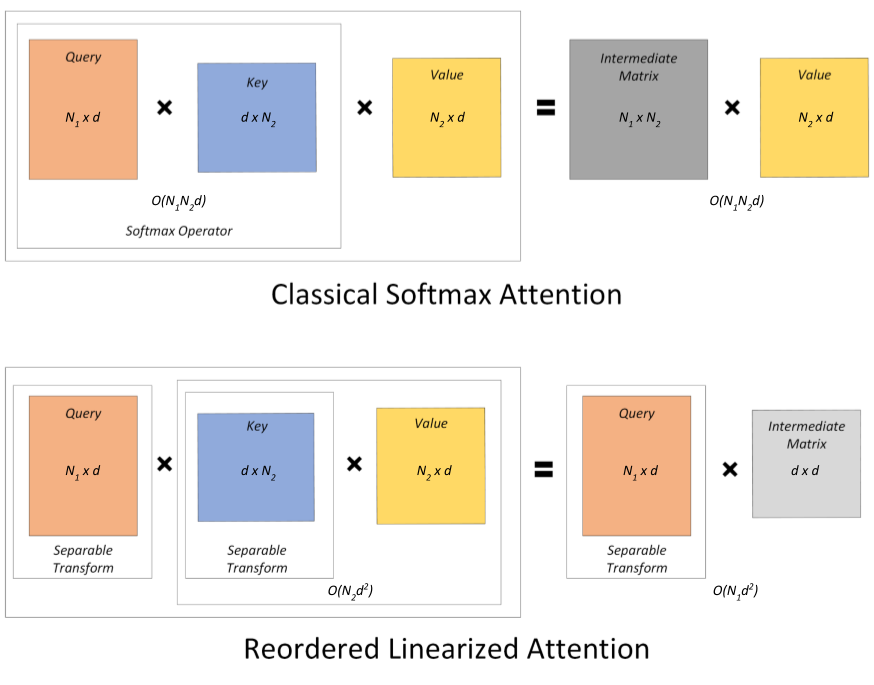}
    \caption{Illustration of attention calculation reordering and linearization for one attention head. This mechanism is useful when both $N_1$ and $N_2$ are significantly larger than $d$, which occurs for many applications or for models with many attention heads with smaller samples. When this condition is met, run-time is linearized from $O(N_1N_2)$ to $O(N_1) + O(N_2)$ for an arbitrary decoding time-step, assuming that neither size is significantly greater than the other.}
    \label{fig:reordering}
    \vspace{-1.5em}
\end{figure}

Katharopoulos et. al \cite{kathrapalous20} introduce the first attention mechanism of truly $O(n)$ run-time complexity with no prior environmental assumptions (e.g. SparseTransformer assumes patterned sparsity exists in the intermediate $QK^T$ matrix) by reordering the quadratic attention calculation, as explained by the caption in Figure \ref{fig:reordering}. This is normally impossible due to the softmax operator removing any associability for the query and key matrices. Katharapoulos et. al elect to replace that softmax operator with a non-linear similarity function that they denote as a function $S$ that is distributable, meaning that $S(Q, K^T) = S_q(Q)S_k(K^T)$. This reordering is demonstrated in a row-wise manner for the output matrix of the attention mechanism via Equations \ref{eq:line_attn}, \ref{eq:redistr_attn}, and \ref{eq:lin_attn} where $A_i$ is equivalent to one output row $i$ in $N_1$ for classical softmax attention and $\Tilde{A}_i$ is equivalent to one output row $i$ in $N_1$ for truly linear attention.

\begin{figure}[h]
    \centering
    \includegraphics[scale=0.35]{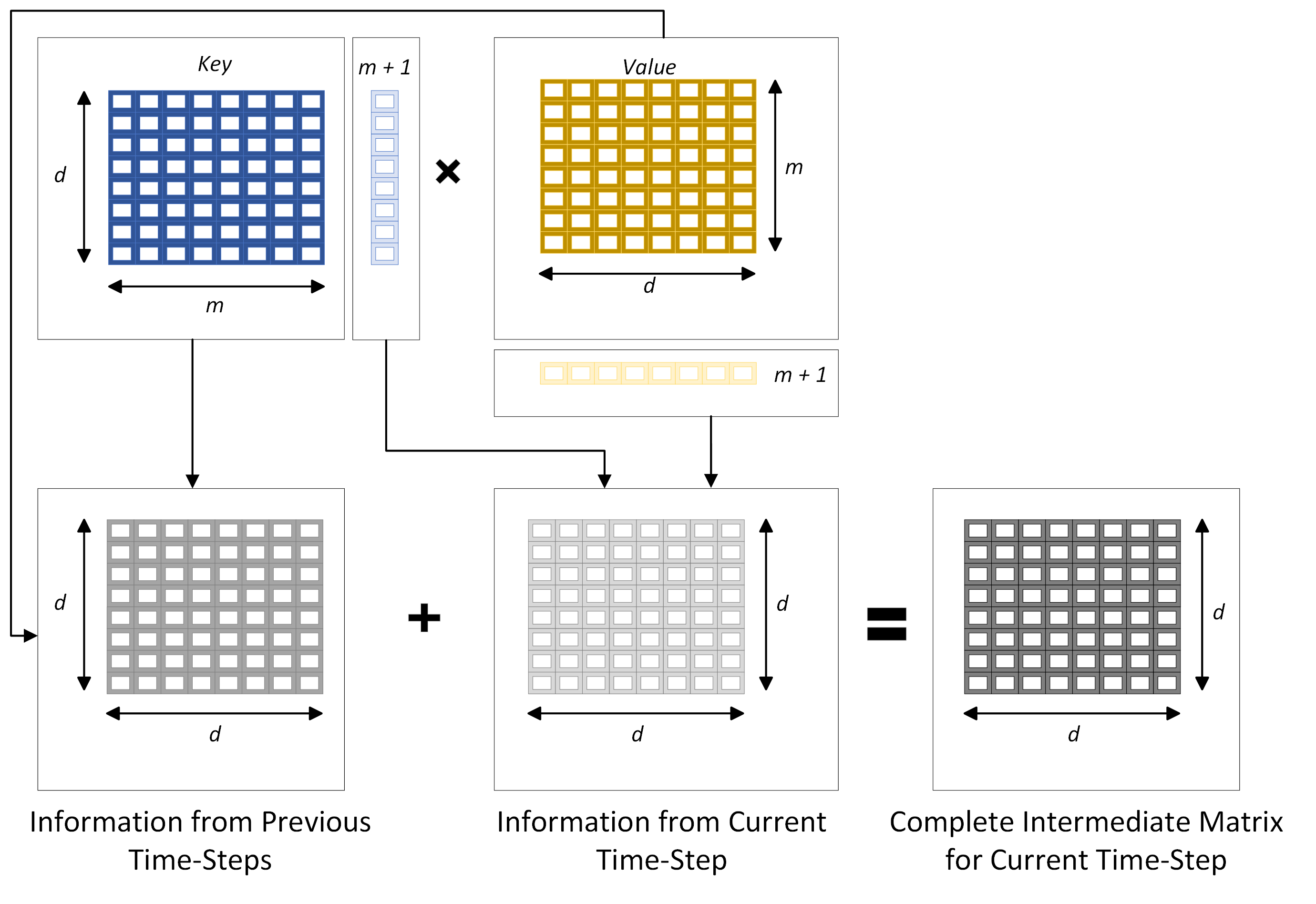}
    \caption{Illustration of one of the data-reuse opportunities present under reordered and linearized attention for the $K^T V$ intermediate matrix for one attention head in decoder self-attention. Past time-steps are represented from decoding time-step 1 to $m$, with the current decoding time-step being represented as $m + 1$.}
    \label{fig:data_reuse}
    \vspace{-2em}
\end{figure}

Critically, Katharapolous et. al note that their implementation can achieve throughput that is orders of magnitude higher than classical attention for very long sequences. This is accomplished for autoregressive tasks via a data-reuse opportunity present in the $K^TV$ intermediate matrix and the normalization vector of a given autoregressive attention calculation, which is highlighted and explained by Figure \ref{fig:data_reuse}.

\vspace{-0.5em}

\begin{equation}
A_i = \sum_j \frac{exp(Q_iK_j^T)}{\sum_jexp(Q_iK_j^T)}V_j
\label{eq:line_attn}
\end{equation}

\vspace{-0.5em}

\begin{equation}
\Tilde{A}_i = \sum_j \frac{S(Q_iK_j^T)}{\sum_jS(Q_iK_j^T)}V_j =  \sum_j \frac{S_q(Q_i)S_k(K_j^T)}{\sum_jS_q(Q_i)S_k(K_j^T)}V_j
\label{eq:redistr_attn}
\end{equation}

\vspace{-0.5em}

\begin{equation}
\Tilde{A}_i = \sum_j \frac{S_q(Q_i)(S_k(K_j^T)V_j)}{S_q(Q_i)\sum_jS_k(K_j^T)}
\label{eq:lin_attn}
\end{equation}

While Katharapolous et. al's work is foundational and inspired the development of a number of linearized attention mechanisms (e.g. \cite{performer2020,zhen2022cosformer}), it can struggle to match the inference quality of other schemes. Their similarity function, defined as $S(M) = ELU(M) + 1$, was settled on seemingly arbitrarily. In response, Choromanski et. al \cite{performer2020} propose Performer, an extremely efficient transformer that makes use of the aforementioned reordering mechanism in addition to several techniques to better emulate softmax-based attention without any prior assumptions. Instead of employing an ELU for their similarity function, Choromanski et. al choose to use positive and orthogonal random feature maps in addition to a customized kernel function to approximate softmax attention. Performer remains a somewhat popular paradigm and has inspired several similar schemes \cite{Ashtari2022FactorizerAS,rfapeng2021,xiong2021nystromformer}. Unfortunately, Performer can suffer from some instability depending on the environment due to its stochastic properties \cite{zhen2022cosformer}.

\vspace{-0.5em}

\subsection{cosFormer and Re-weighting Mechanisms}
Similar to its prominent predecessors, cosFormer, proposed by Qin et. al \cite{zhen2022cosformer}, takes advantage of the reordering opportunities present in linearized attention when the softmax operator is replaced with an approximate similarity function. Additionally, a re-weighting mechanism is proposed that emphasizes the locality bias (i.e. a tendency for nearby tokens to attend strongly to one another) observed in the intermediate $QK^T$ matrix of many NLP tasks \cite{bertlooksat2019,localitybert2019} and ensures the positivity of the $QK^T$ matrix. This allows for tight approximation of the softmax operator (i.e. probability distribution concentration, positive $QK^T$ for every score) in linear time and without the random characteristics of Performer. Given that, they employed the decomposable re-weighting function described in Equation \ref{eq:cos_attn} to self-attention blocks, where $N$ is the length of the input sequence and $S_q$ and $S_k$ are ReLU functions. The maximal response of the function in Equation \ref{eq:cos_attn} occurs when the difference between positions in the query and key are equal and tapers off as that positional difference grows, modeling softmax behavior for attention mechanisms exhibiting locality bias. 


\begin{equation}
S(Q_i, K_j^T) = S_q(Q_i)S_k(K_j^T)cos(\frac{\pi}{2}(\frac{i-j}{N}))
\label{eq:cos_attn}
\end{equation}


cosFormer achieves state-of-the-art scores for a number of tasks, including fixed-length autoregressive language modeling on WikiText-103 \cite{baevski2019adaptive}, bi-directional language modeling via insertion into RoBERTa \cite{liu2019roberta} during pre-training, downstream fine-tuning for various text classification tasks on RoBERTa, and competitive results on the Long-Range Arena (LRA) benchmark \cite{tay2021long}. The LRA scores are especially superb given the locality-based assumptions baked into cosFormer's re-weighting mechanism that should discourage long-range dependencies. 




\subsection{Motivation for Investigation}
While many capable efficient attention mechanisms exist, their widespread applicability is usually limited by prior assumptions (e.g. assuming sparsity or expressivity of low-rank approximations) or by poor inference quality outside of a few tasks. Even for a single application, multiple attention blocks can behave significantly differently such that broadly applying a single efficient attention mechanism may produce poor results. In spite of these difficulties, many complex autoregressive applications such as S2T NMT, SimulST, and TTS, desire efficient attention mechanisms for their associated throughput increases. 

Applying a state-of-the-art technique like cosFormer to mitigate inference quality loss for the aforementioned applications is natural. However, some issues exist for cosFormer when it is applied to autoregressive tasks without a fixed length. First and foremost, cosFormer has no method to deal with the mismatch between casual training and the evaluation environment as far as target sequence length availability is concerned. Simply stepping the target sequence length at each decoding time-step (i.e. setting it equal to the current sequence length) is a poor solution for most tasks and leads to unstoppable inference for TTS with absurd distortion and BLEU scores of essentially 0 for en-de NMT and SimulST. Moreover, while locality bias is often present for attention mechanisms, it is not always relevant depending on the application and on the attention mechanism. We will demonstrate later in this paper that applications often do not meet expectations about their emphasis or lack thereof on token locality for various attention blocks. Additionally, cosFormer was never extended to decoder cross-attention, necessarily limiting the scope of its applicability. 

Others have attempted to apply cosFormer to new tasks, but have applied this linearization method to autoregressive tasks, in particular, with extremely limited success. Liu et. al \cite{liu2022neural} worked to apply cosFormer to text-to-text en-de NMT as part of an architecture search enhancement \cite{hu2021ranknas}. However, linearizing any of the decoder attention blocks in their study resulted in BLEU scores of 0. Additionally, as they explored the validity of applying cosFormer to new tasks, they varied a number of parameters tied to the models they studied, including number of layers, FFN dimensions, and model embedding dimensions, rendering their evaluation of cosFormer as a methodology somewhat unreliable. 

\vspace{-0.5em}

\section{Proposed Approach}
\subsection{Modular Linearized Attention}
In recognizing that cosFormer, and other linearization techniques, are often built with specific properties in mind that are not always present on an application-basis, we propose \textit{modular linearized attention (MLA)}. This method combines efficient attention mechanisms in a modular manner to take advantage of the strengths of different linearization schemes while avoiding applying them to attention blocks that do not exhibit desired characteristics. This paper firmly urges model designers to avoid seeking a ``one size fits all'' attention linearization solution, as we will later demonstrate that these solutions are typically  inferior to a modular combination in terms of inference quality. We provide a brief formulation of our method via Equation \ref{eq:search_high_level} for an encoder-decoder transformer-based model $T$.

\vspace{-1em}

\begin{equation}
T(eSA(A_t), \: dSA(A_t), \: dCA(A_t)), \: \: A_t = \{softmax, \: cosFormer, \: ReLU\}
\label{eq:search_high_level}
\end{equation}

As described in Equation \ref{eq:search_high_level}, our total search space is denoted as softmax, cosFormer, or simple ReLU attention (identity re-weighting function) for the following attention blocks: $eSA$ for encoder self-attention, $dSA$ for decoder self-attention, and $dCA$ for decoder cross-attention. To determine an appropriate combination of attention blocks for each task, we propose targeted ablation studies on an attention block-basis. It should be noted that a lower level of granularity could be applied for layer-by-layer \textit{MLA} (i.e. applying cosFormer to multiple layers in the decoder while applying simple ReLU to layers with attention blocks that do not emphasize token locality), but we leave this to future work to explore.

\vspace{-0.5em}

\subsection{Augmenting cosFormer for Decoder Cross-Attention}
In the interest of enabling \textit{MLA} for state-of-the-art linearization techniques that employ sequence length-based re-weighting, in the vein of cosFormer, we propose a generalization of cosFormer for decoder cross-attention. This generalization is defined by Equation \ref{eq:cos_gen_attn}, which preserves cosFormer's original design intent behind locality between relative positions by applying them to both sequences being attended to in cross-attention ($M$ is set equal to the length of the key and value matrices). Failing to produce this generalization, as in the original cosFormer, would preclude investigation of cosFormer for many possible attention mechanisms, limiting the modular attention search space unnecessarily. 

\begin{equation}
S(Q_i, K_j^T) = S_q(Q_i)S_k(K_j^T)cos(\frac{\pi}{2}(\frac{i}{N} - \frac{j}{M}))
\label{eq:cos_gen_attn}
\end{equation}

\vspace{-1em}

\subsection{Closing the Gap: Target Sequence Length Prediction}
Any sequence length-based re-weighting mechanism, like the one cosFormer employs, suffers from a mismatch between the training and evaluation environments concerning the availability of the target sequence length in autoregressive tasks. For autoregressive tasks without a fixed sequence length, the target sequence is unbounded and will only cease after an end of sequence prediction. Two obvious paths to close that gap present themselves. One option would be to replicate the stepping behavior of the current sequence length in the re-weighting mechanism at every decoding time-step (i.e. increasing $N$ and/or $M$ in Equation \ref{eq:cos_gen_attn}). This adds a number of sequential steps that, in practice, slow down training significantly or introduce massive memory footprints to enable parallelization of those sequential steps. Moreover, this does not reflect the set history of autoregressive applications. For example, during decoder self-attention, the key entry for time-step $m$ is constant for all later time-steps in softmax-based attention, but would differ at later time-steps for this stepping scheme, which is intuitively erroneous. Alternatively, one could attempt to predict the target sequence length in the evaluation environment, and we consider this option to better reflect autoregressive behavior. 


This paper proposes a few simple and computationally cheap solutions for target sequence length prediction. The first is a statistical analysis of the training set, producing a simple ratio $\alpha$ to map between the average source length and target length. The second is a lookup table based on a similar statistical analysis on the average length mapping on a token-to-token basis. The third is learned target length prediction via a compact network of a few convolutional layers and ReLU activations. We validate each of these approaches briefly in the following sections on TTS with a model based on TransformerTTS \cite{ttstransformer} and reduced dimensionality (training and evaluation details are elaborated upon in later sections), with preliminary results for the translation tasks suggesting that observed trends for TTS extend to translation. Ultimately, we conclude that a simple ratio $\alpha$ provides competitive results while remaining computationally negligible, and we choose to employ it for our evaluation in later sections. Mel-cepstral distortion (MCD) is the quantitative metric of choice for this exploration, as it isolates phonetic differences between the synthesized and reference spectrograms \cite{Kubichek1993MelcepstralDM,prosody2018,fairseq_s2_2021}.

\vspace{-1em}

\subsubsection{Sequence Length Prediction via Simple Ratio $\alpha$}
A brief analysis of the training set of LJSpeech \cite{ljspeech17} suggests a ratio $\alpha$ of around 1.25 between the source sequence to the target sequence for average mapping and a ratio $\alpha$ of around 1.5 for close to 90\% of training examples containing positive queries and keys in Equation \ref{eq:cos_gen_attn}. The primary question facing this method of target length prediction is clear: is mostly guaranteeing a positive query and key matrix (ensured in softmax-based attention) actually as beneficial cosFormer's construction would suggest? As Table \ref{tab:mcd_ratio_cross} demonstrates, this is not necessarily the case. For example, an $\alpha$ of 1.75 would almost always guarantee positive query and key matrices, but the distortion per reference frame in Table \ref{tab:mcd_ratio_cross} is very large and the summed synthesized sequence lengths are nearly 2.5x the summed reference sequence lengths. Given that, we default to average mappings for later evaluations. We note that this method is computationally negligible during evaluation, and fits well into very tight computational budgets. 

\begin{table}[h]
    \centering
    \begin{tabular}{c|c|c|c}
        $\alpha$ Length Ratio & Synth. Dur. (frames) & Dist./Ref. & Dist./Align. \\
        \hline
         0.50 & 167k & 7.42 & 7.05 \\
         \hline
         0.60 & 212k & 7.73 & 6.99 \\
         \hline
         0.75 & 307k & 8.49 & 6.59 \\
         \hline
         1.00 & 431k & 9.70 & 5.75 \\
         \hline
         1.25 & 531k & 10.74 & 5.27 \\
         \hline
         1.50 & 650k & 12.00 & 4.88 \\
         \hline
         1.75 & 772k & 13.30 & 4.59
    \end{tabular}
    \caption{MCD values (lower is better) for differing target length to source length ratios for target length predictions. Synthesized duration in frames for all samples is supplied to showcase generation tendencies for each $\alpha$. All models were baseline TransformerTTS architectures with their decoder cross-attention mechanism replaced by cosFormer. Reference duration is approximately 273k frames across all 523 test samples in LJSpeech. Distortion is provided as average distortion per reference frame and per dynamic time-aligned frame \cite{berndt1994,wavetac2020}, as optimizing for one or the other would lead to clearly erroneous $\alpha$ values, as demonstrated in this table.}
    \label{tab:mcd_ratio_cross}
    \vspace{-3em}
\end{table}

\vspace{-1em}

\subsubsection{Sequence Length Prediction via Lookup Table}
The lookup table (LUT) is constructed as a set of mappings from input tokens to the average size of their representation in the output sequence via a statistical analysis similar to the one done for a simple ratio $\alpha$. This method tends to excel when the vocabulary size is relatively small and when token mappings are consistent. The former characteristic minimizes the memory footprint of the LUT and ensures a large enough set of samples for each token mapping (unknown tokens used the average token-to-token mapping) while the latter ensures accurate mappings. In the case of the TTS task exemplified in this section, we pre-processed the input text into phonemes \cite{fairseq_s2_2021} and produced reference alignments via the Montreal Forced Aligner (MFA) \cite{MontrealFA2017}. Phonemes are well-suited to this methodology as they should have consistent mappings from the source to target sequence that can be taken advantage of and form a tight vocabulary. In practice, as shown in Table \ref{tab:mcd_lut}, this method performed similarly to a simple ratio $\alpha$, and we attribute this unexpected lack of improvement to extreme variance in predicting the length of silences (or ``space'' phonemes), errors introduced by MFA itself, and that this method does not take advantage of sequence context. A dampening factor that scales the average mappings is briefly explored as a fine-tuning tool to discourage or encourage the provided model to produce an end of sequence prediction.

\begin{table}[h]
    \vspace{-2em}
    \centering
    \begin{tabular}{l|c|c|c}
        Dampening Factor & Synth. Dur. (frames) & Dist./Ref. & Dist./Align. \\
        \hline
        1.1 & 342k & 8.81 & 6.11 \\
        \hline
        1.0 & 378k & 9.20 & 5.89 \\
        \hline
        0.9 & 450k & 9.56 & 5.55 \\
    \end{tabular}
    \caption{MCD values for various LUT dampening factors. All models contained cosFormer attention blocks for decoder cross-attention with softmax self-attention blocks.}
    \label{tab:mcd_lut}
    \vspace{-5em}
\end{table}

\subsubsection{Sequence Length Prediction via Compact Neural Network}
An intuitive solution, training a compact network to predict the target sequence length should acknowledge source sequence context, unlike the other evaluated methodologies, and result in higher quality predictions. The baseline TransformerTTS was augmented with this network (inspired by the very similar network in FastSpeech2 \cite{fastspeech2}, a non-autoregressive TTS solution) at the end of the encoder stack, and for the following experiment the target sequence length predictor was composed of two convolutional layers with a ReLU activation function tied to the end of each of them. Importantly, limiting the size of this network is required for a fair comparison, as if it requires notable computation time and significantly increases the memory footprint of the model, then it is likely unsuitable for applications where linearized attention is desirable in the first place. That limited size necessarily degrades the quality of the target length prediction and, as seen in Table \ref{tab:mcd_learned_tgt_pred}, results in inference quality that is just competitive with the previous two methodologies. We leave it to future work to more closely examine the performance to efficiency trade-off for this augmentation. 

\begin{table}[h]
    \centering
    \begin{tabular}{l|c|c|c}
        Linearization Scheme & Synth. Dur. (frames) & Dist./Ref. & Dist./Align. \\
        \hline
        cosFormer $dSA$ \& $dCA$ & 328k & 9.09 & 6.68 \\
        \hline
        cosFormer $dSA$ & 334k & 8.98 & 6.58 \\
        \hline
        cosFormer $dCA$ & 367k & 9.38 & 6.35 \\
    \end{tabular}
    \caption{MCD values for various cosFormer ablations with learned target length prediction. Ablations made use of softmax for all attention blocks other than the specified linear ones.}
    \label{tab:mcd_learned_tgt_pred}
    \vspace{-1.5em}
\end{table}


\section{Training and Evaluation Details}
To demonstrate the effectiveness of \textit{MLA} we formalize the following experiments on Fairseq \cite{fairseq2019}, a language and sequence-modeling toolkit built in PyTorch. TTS experiments employed the Fairseq $S^2$ extension \cite{fairseq_s2_2021}, which contains a number of useful tools for pre-processing and evaluation. TTS was thoroughly explored while additional experiments were also conducted for NMT and SimulST tasks.

\subsection{Model Configurations and Training Hyperparameters}
 All models trained for the following TTS experiments were based on TransformerTTS \cite{ttstransformer} with a length predictor augmentation based on the aforementioned schemes. The S2T NMT and SimulST models were based on models within the ESPnet-ST toolkit that focused on end-to-end S2T translation with a modified cross-attention block for a wait-k and fixed pre-decision paradigm \cite{mma19,mono_xma20,simulst_xma20}. To avoid cascading multiple models \cite{bentivogli-etal-2021-cascade}, all translation models were pre-trained for automatic speech recognition (ASR) and their encoders were used for initialization when training for en-de S2T NMT. The models employed in this paper are nearly identical to these baselines, but with reduced dimensionality to respect the expected resource-constraints for environments where efficient attention is often employed. All models were trained on four NVIDIA Tesla V100 GPUs. Further details for model parameters and training hyperparameters can be found in the supplementary materials.

\vspace{-1em}

\subsubsection{End of Sequence Training for TTS}
End of sequence prediction is engaged with separately for TTS and is a somewhat sensitive part of the training process, with a single positive output resulting in an end of sequence prediction. If the end of sequence linear layer converges to a poor solution, unstoppable inference can occur and significant overgeneration on the order of 3x to 4x the number of reference output frames can be observed. To compensate for this issue, a recommended positive weighting of 5.0 in Wang et. al's work \cite{fairseq_s2_2021} was applied to a separately calculated end of sequence loss that serves as an additional training objective. 

\vspace{-1em}

\subsubsection{Target Length Prediction Training}
For learned target length prediction, we adopt a very similar approach to FastSpeech2 \cite{fastspeech2}. Training goals related to target length prediction were added via separately calculated loss values for the predicted target length on a phoneme-by-phoneme basis, generated via the Montreal Forced Aligner \cite{MontrealFA2017}. Target sequence length prediction (and cosFormer) was not applied to simultaneous translation tasks, as predicting the target length is not possible without the oracle knowledge on when a speaker would stop.

\vspace{-0.5em}

\subsection{Evaluation Setup and Metrics}
All models were evaluated within Fairseq's framework or with associated extensions. For SimulST, we employed SimulEval \cite{simuleval2020} as an evaluation framework and evaluated on a wait-k of 3 and a fixed pre-decision ratio of 7 \cite{simulst_xma20} (we noted quality improvements when training on a wait-k of 5 and pre-decision ratio of 9, despite the mismatch). All TTS evaluations were executed on a single NVIDIA Tesla V100 and all translation evaluations were executed on a Intel Xeon Platinum 8168 CPU.

\vspace{-1em}

\subsubsection{Distortion (MCD) and BLEU}
Mel cepstral distortion (MCD) \cite{Kubichek1993MelcepstralDM} and mel spectral distortion (MSD) have emerged as popular quantitative metrics for speech synthesis \cite{prosody2018,fairseq_s2_2021,wavetac2020}. MCD is calculated as demonstrated in Equation \ref{eq:mcd} with dimensionality $K$ of 13 and with the Mel-Frequency Cepstrum Coefficients (MFCCs) $c_{i,k}$ and their synthesized counterparts being calculated via classical methods. MSD is calculated similarly, but contains slightly different embedded information. For the sake of brevity, only MCD is included in later sections.

\begin{equation}
    MCD_K = \frac{1}{T}\sum_{t=0}^T\sqrt{\sum_{k=1}^K(c_{i,k} - c'_{i,k})}
    \label{eq:mcd}
\end{equation}

All translation models were evaluated via 4-gram BLEU on sacreBLEU \cite{sacrebleu2018}. To ensure a fair comparison to other, similar models, all translations were detokenized before scoring.  

\section{Results}
\subsection{TTS Training Results for Targeted Ablations}
To determine how well individual attention blocks demonstrate desired characteristics for each of the attention mechanisms in our proposed search space, we refer to the training results in Table \ref{tab:baselines} for targeted ablations. cosFormer encoder self-attention and decoder cross-attention ablations perform particularly well. However, a simple ReLU similarity function outperforms cosFormer for decoder self-attention. We note that the decoder self-attention ReLU ablation has slightly higher composite loss than full cosFormer, but combining it with cosFormer via \textit{MLA} should eliminate the EOS loss entirely, as any cosFormer decoder attention block perfectly predicts the end of sequences during training.

It was assumed that TTS, as an application, would exhibit a focus on token locality for all attention blocks, but these training results indicate otherwise for decoder self-attention and validate the necessity of an identity re-weighting mechanism as a performance baseline for this design search. It is worth noting that cosFormer ablations generally overestimate their performance when they are embedded within the decoder, as during training (unless otherwise stated) target sequence lengths are known. We underscore the very poor training results of fully linearized models with homogenous attention linearization schemes, demonstrating the potential of \textit{MLA}. 


\begin{table}[h]
    \vspace{-1.5em}
    \centering
    \begin{tabular}{l|c|c|c}
        Linearization Scheme & Composite Loss & MSE Loss & EOS Loss \\
        \hline
        Softmax Attention (non-Linear) & 1.021 & 0.364 & 0.026 \\
        \hline
        Full cosFormer & 1.132 & 0.440 & 0.000 \\
        \hline
        cosFormer $eSA$, Softmax $dSA$ \& $dCA$ & 1.063 & 0.385 & 0.029 \\
        \hline
        cosFormer $dSA$, Softmax $eSA$ \& $dCA$ & 1.158 & 0.455 & 0.000 \\
        \hline
        cosFormer $dCA$, Softmax $eSA$ \& $dSA$ & 1.102 & 0.424 & 0.000 \\
        \hline
        Full Simple ReLU & 1.262 & 0.496 & 0.029 \\
        \hline
        Simple ReLU $eSA$, Softmax $dSA$ \& $dCA$ & 1.097 & 0.399 & 0.037 \\
        \hline
        Simple ReLU $dSA$, Softmax $eSA$ \& $dCA$ & 1.133 & 0.424 & 0.025 \\
        \hline
        Simple ReLU $dCA$, Softmax $eSA$ \& $dSA$ & 1.162 & 0.438 & 0.031 \\
    \end{tabular}
    \caption{TTS training results for various linearization schemes and their ablations. All loss values (lower is better) are provided from the best performing checkpoint on the validation split. Composite loss encompasses all training goals, including MAE loss, MSE loss, and EOS loss. EOS loss corresponds to the separately trained end of sequence loss, which is a critical indicator of inference quality. MSE loss is included explicitly as it served as a tie-breaker for checkpoints with otherwise nearly identical training performance.}
    \label{tab:baselines}
    \vspace{-4.5em}
\end{table}


\subsection{Training and Evaluation Results for Finalized TTS Configurations}
Based on the training results in Table \ref{tab:baselines}, we propose a few  \textit{MLA} configurations with optimally placed efficient attention mechanisms for this search. The most efficient combination includes cosFormer being applied to encoder self-attention and decoder cross-attention, with decoder self-attention employing a simple ReLU baseline. However, linearizing encoder self-attention and decoder self-attention, while clearly benefiting calculated FLOPs in Table \ref{tab:latency}, does not influence the practical run-time of the model as significantly. As such, a final model configuration with notable throughput increases while also mitigating inference performance degradation would likely only host cosFormer in the decoder cross-attention block. Direct comparisons to the ablations in Table \ref{tab:baselines} with the final configurations in Table \ref{tab:final_train} demonstrate the effectiveness of \textit{MLA} in producing competitive models, especially noting that the fully linearized \textit{MLA} model produces better composite loss on the validation set (a difference of 0.013 composite loss) when compared to a model with all attention blocks replaced with cosFormer.

All decoder cosFormer attention blocks made use of a simple ratio $\alpha$ of 1.25 during evaluation. Table \ref{tab:fin_pred_qual_mcd} represents the distortion of these final configurations. While some additional distortion is present in spite of the application of \textit{MLA} (1.61 added average distortion per dynamic time-aligned frame), this distortion is mitigated for this search space while still achieving notable throughput increases. As can be observed in Table \ref{tab:latency}, around a 5.2\% decoder throughput increase (decoder run-time tends to dominate encoder run-time for these tasks, SimulST excepted) was produced with mitigated distortion while the largest decoder speedup produced was around 7\% with a fully linearized model, demonstrating the capability of \textit{MLA} to produce significant efficiency gains with relatively small added distortion.

\begin{table}[h]
    \vspace{-1.5em}
    \centering
    \begin{tabular}{l|c|c|c}
        Linearization Scheme & Composite Loss & MSE Loss & EOS Loss \\
        \hline
        Softmax Attention (non-Linear) & 1.021 & 0.364 & 0.026 \\
        \hline
        cosFormer $eSA$ \& $dCA$, ReLU $dSA$ & 1.119 & 0.433 & 0.000 \\
        \hline
        cosFormer $eSA$ \& $dCA$, Softmax $dSA$ & 1.105 & 0.426 & 0.000 \\
        \hline
        cosFormer $dCA$, Softmax $eSA$ \& $dSA$ & 1.102 & 0.424 & 0.000 \\
    \end{tabular}
    \caption{TTS training results for final \textit{MLA} models with best performing checkpoints on the validation set listed. Softmax is provided again for a quick reference point.}
    \label{tab:final_train}
    \vspace{-3.5em}
\end{table}

\begin{table}[h]
    \vspace{-1.5em}
    \centering
    \begin{tabular}{l|c|c|c}
        Linearization Scheme & Synth. Dur. (frames) & Dist./Ref. & Dist./Align. \\
        \hline
        Softmax Attention (non-Linear) & 285k & 5.52 & 4.69 \\
        \hline
        cosFormer $eSA$ \& $dCA$, ReLU $dSA$ & 273k & 8.10 & 6.57 \\
        \hline
        cosFormer $eSA$ \& $dCA$, Softmax $dSA$ & 293k & 8.35 & 6.43 \\
        \hline
        cosFormer $dCA$, Softmax $eSA$ \& $dSA$ & 297k & 8.20 & 6.30 \\
    \end{tabular}
    \caption{MCD values for various final \textit{MLA} configurations, two of which are not fully linearized due to the size of this workload warranting comparisons with non-linear ablations.}
    \label{tab:fin_pred_qual_mcd}
    \vspace{-4em}
\end{table}

\begin{table}[h]
    \vspace{-1em}
    \centering
    \begin{tabular}{l|c|c|c}
        Linearization Scheme & Enc. Thrpt. (itr/sec) & Dec. Thrpt. (itr/sec) & FLOPs\\
        \hline
        Softmax Attention (non-Linear) & 1.51 & 72.81 & 1.94G \\
        \hline
        cosFormer $eSA$ \& $dCA$, ReLU $dSA$ & 1.57 & 77.88 & 1.46G \\
        \hline
        cosFormer $eSA$ \& $dCA$, Softmax $dSA$ & 1.58 & 76.53 & 1.60G \\
        \hline
        cosFormer $dCA$, Softmax $eSA$ \& $dSA$ & 1.52 & 76.56 & 1.68G \\
    \end{tabular}
    \caption{Efficiency related results from various \textit{MLA} configurations generating spectrograms on the LJSpeech test set. Encoder and decoder throughput (higher is better) was measured via the number of forward calls and the wall-clock time for those calls. FLOPs (lower is better) were calculated and are an estimation of floating point operations (all operations treated as equal) for a single sample with a source sequence length of 100 and a target sequence length of 150. Small variations in throughput are attributed to slightly different device conditions between efficiency tests.}
    \label{tab:latency}
    \vspace{-3.5em}
\end{table}

\subsection{en-de NMT and SimulST Training and Evaluation Results}
Both en-de S2T NMT and SimulST were explored less exhaustively than TTS. We leave it open to future work to elaborate on \textit{MLA} for these tasks in addition to exploring target length prediction schemes for simultaneous tasks such that length-based re-weighting mechanisms can be applied during attention linearization. For all cosFormer blocks, the target length was predicted via a simple ratio $\alpha$ of 0.6. Concerning NMT, results can be observed in Tables \ref{tab:nmt_ablation} and \ref{tab:224nmt_ablation}. While homogeneous, fully linearized models suffered from poor accuracy, models featuring modular linearization and including some quadratic elements tended to exhibit mitigated performance degradation, or even competitive prediction quality, compared to fully quadratic models. In particular, we note that a simple ReLU baseline for $dSA$ performed competitively with a softmax-based implementation for both S2T NMT and SimulST, exhibiting gains of 0.6 BLEU and 0.44 perplexity respectively.

\begin{table}[!h]
    \centering
    \begin{tabular}{l|c|c}
         Attention Linearization Scheme & BLEU & ppl(dev) \\
         \hline
         Softmax Attention (non-Linear) & 10.47 & 9.36 \\
         \hline
         Full cosFormer & 2.37 & 20.08 \\
         \hline
         Full Simple ReLU & 1.95 &  18.33 \\
         \hline
         cosFormer $dSA$, Softmax Elsew. & -- & 9.92 \\
         \hline
         Simple ReLU $dSA$, Softmax Elsew. & 11.07 & 9.74 \\
    \end{tabular}
    \caption{Results from S2T NMT for MuST-C en-de for various linearization schemes with softmax as a baseline. BLEU scores (higher is better) are generated at inference on the test set, are detokenized, and are generated via sacreBLEU. Perplexity (lower is better) is generated during training on the validation set. Missing entries were not explored due to resource/time constraints.}
    \label{tab:nmt_ablation}
    \vspace{-2em}
\end{table}

\begin{table}[!h]
    \centering
    \begin{tabular}{l|c|c}
         Attention Linearization Scheme & BLEU & ppl(dev) \\
         \hline
         Full cosFormer & 2.73 & 22.12 \\
         \hline
         cosFormer $eSA$ \& $dSA$, Simple ReLU $dCA$ & 1.89 & 23.1 \\
         \hline
         cosFormer $eSA$ \& $dSA$, Softmax $dCA$ & 7.71 & 13.35 \\
    \end{tabular}
    \caption{Results from S2T NMT for MuST-C en-de for various linearization schemes with slightly different model parameters. These models were trained with an embedding dimension $d_{model}$ of 224 and 4 attention heads.}
    \label{tab:224nmt_ablation}
    \vspace{-2em}
\end{table}

\begin{table}[!h]
    \centering
    \begin{tabular}{l|c|c}
         Attention Linearization Scheme & BLEU & ppl(dev) \\
         \hline
         Softmax Attention (non-Linear) & 9.25 & 10.15 \\
         \hline
         Simple ReLU $eSA$ \& $dSA$, Softmax $dCA$ & 7.35 & 12.81 \\
         \hline
         Simple $dSA$, Softmax $eSA$ \& $dCA$ & -- & 9.71 \\
    \end{tabular}
   \caption{Results from SimulST for MuST-C en-de for various linearization schemes. Missing entries were not explored due to resource/time constraints.}
    \label{tab5:simulst_ablation}
    \vspace{-3em}
\end{table}

We note that cosFormer's superior cross-attention performance during training and evaluation worked in opposition to expectations. While English is a subject-verb-object (SVO) language, German is a subject-object-verb (SOV) language, implying that long-distance dependencies should exist for this tasks's cross-attention block. This validates the necessity of a targeted ablation study to empirically show the capabilities of a given linearization scheme for each attention block.



\vspace{-0.75em}

\section{Conclusion}
Attention linearization is a popular method for building efficient transformers for a variety of latency-sensitive tasks. In this paper, we propose \textit{modular linearized attention (MLA)}, which combines multiple efficient attention mechanisms to maximize performance. To enable \textit{MLA} and solve issues in applying cosFormer \cite{zhen2022cosformer} to autoregressive tasks without a fixed length, we propose several computationally cheap solutions for target sequence length prediction. Moreover, we apply effective attention linearization with sequence length-based re-weighting for the first time to S2T NMT, SimulST, and autoregressive TTS. We achieve up to a 7\% decoder throughput increase for TTS with mitigated additional MCD and produce competitive prediction quality for en-de S2T NMT and SimulST, with a up to a 0.6 BLEU score increase and 0.44 perplexity reduction for models featuring \textit{MLA}.

\bibliographystyle{splncs04}
\bibliography{bibliography}

\newpage

\section{Supplementary Materials}
\subsection{Codebase and Implementation}
The codebase for this paper will be released upon its acceptance.

\subsection{Expanded Model Parameters}
We expand upon the model parameters briefly described in this paper below. Parameters not listed can be assumed to be identical to vanilla implementations of these models. 

\subsubsection{TTS Transformer Parameters}
\begin{itemize}
    \item Encoder Layers: 6
    \item Decoder Layers: 6
    \item Embedding Dim. $d_{model}$: 256
    \item Attention Heads: 8
    \item FFN Hidden Dim. $d_{ffn}$: 1024
    \item Conv. Pre-net Layers: 4
    \item Conv. Pre-net Kernel Size: 4
    \item Conv. Post-net Layers: 4
    \item Conv. Post-net Kernel Size: 4
\end{itemize}

\subsubsection{ESPNet-ST Model Parameters}
\begin{itemize}
    \item Encoder Layers: 6
    \item Decoder Layers: 6
    \item Embedding Dim. $d_{model}$: 256
    \item Attention Heads: 8
    \item FFN Hidden Dim. $d_{ffn}$: 1024
    \item Conv. Pre-net Layers: 4
    \item Conv. Pre-net Kernel Size: 4
\end{itemize}


\subsection{Training Hyperparameters}
All TTS models were trained with Adam \cite{adam2014} as an optimizer with typical parameters: $\beta_1$ and $\beta_2$ were set to 0.9 and 0.98 respectively, the learning rate was set to 2e-3, and the learning rate scheduler used an inverse square-root to decay the learning rate. The models were trained with dynamic batching, warmed up for 4000 updates, and trained for around 18000 updates in total with gradients that were clipped to 5.0 and layer and attention dropouts of 0.1. TTS models were trained on the single-speaker, English LJSpeech dataset \cite{ljspeech17} with recommended splits for training, validation, and evaluation at a sampling frequency of 16 kHz. 

The NMT and SimulST tasks were trained on the en-de version of the Must-C dataset \cite{mustc2021}, composed of several hundred hours of audio sampled at 22 kHz and with similarly recommended splits for training, validation, and evaluation. For the translation tasks, all models were also optimized via Adam with similar parameters and a learning rate set to 6e-4 with an identical learning rate scheduler. The models were trained with dynamic batching, warmed up for 8000 updates, starting with a learning rate of 1e-4, and trained for around 14000 updates in total with gradients clipped to 10.0 on the MuST-C English to German dataset \cite{mustc2021}. SimulST models were trained with a wait-k of 5 and pre-decision ratio of 9.

\subsection{Addressing Causal Attention Linearization During Training}
We briefly explored applying \textit{MLA} to increase training speed for causal attention, but linear decoder self-attention is inefficient without specialized CUDA implementations \cite{kathrapalous20} (also applies to decoder cross-attention for simultaneous applications). This is due to the fact that the space complexity of a naive approach for linearized attention during training for decoder self-attention, for example, would require $O(nd^2)$ space complexity, as opposed to the $O(d^2)$ space complexity required during inference, as distinct $K^TV$ matrices are required for every decoding time-step. To compensate, a reduction in batch size is often necessary, but, in practice, this results in overall slower training. We note that this is an issue for any truly linear attention calculation being applied to causal attention, and note that the solutions proposed in this paper are focused on improving test-time performance. Given this paper's focus on optimization during deployment, we leave it to future work to solve this problem for efficient linearized training (cosFormer's \cite{zhen2022cosformer} codebase did not address this issue either).

\subsection{Brief Analysis of Expanded Network for Learned Target Length Prediction}
We briefly explored expanding the size of the compact network for learned target length prediction in this paper, but found that the network in the main body of this paper is roughly appropriately sized, all other facets of this model and tasks being held equal. In fact, as demonstrated in Table \ref{tab:mcd_learned_tgt_pred_exp}, comparable inference quality was observed for a breadth expansion (increased hidden dimension size) and inference quality loss was observed for a depth expansion (increased number of layers). 

\begin{table}[h]
    \centering
    \begin{tabular}{l|c|c|c}
        Learned Length Predictor Size & Synth. Dur. (frames) & Dist./Ref. & Dist./Align. \\
        \hline
        Baseline & 367k & 9.38 & 6.35 \\
        \hline
        Doubled Hidden Dim. & 314k & 8.72 & 6.53 \\
        \hline
        Additional Conv. Layer & 436k & 10.51 & 5.96 \\
    \end{tabular}
    \caption{MCD values for various cosFormer ablations with learned target length prediction. Ablations made use of softmax for all attention blocks other than the specified linear ones.}
    \label{tab:mcd_learned_tgt_pred_exp}
    \vspace{-1.5em}
\end{table}

\newpage

\subsection{Additional Transformer Illustrations}
\begin{figure}[!htb]
    \centering
    \includegraphics[scale=0.6]{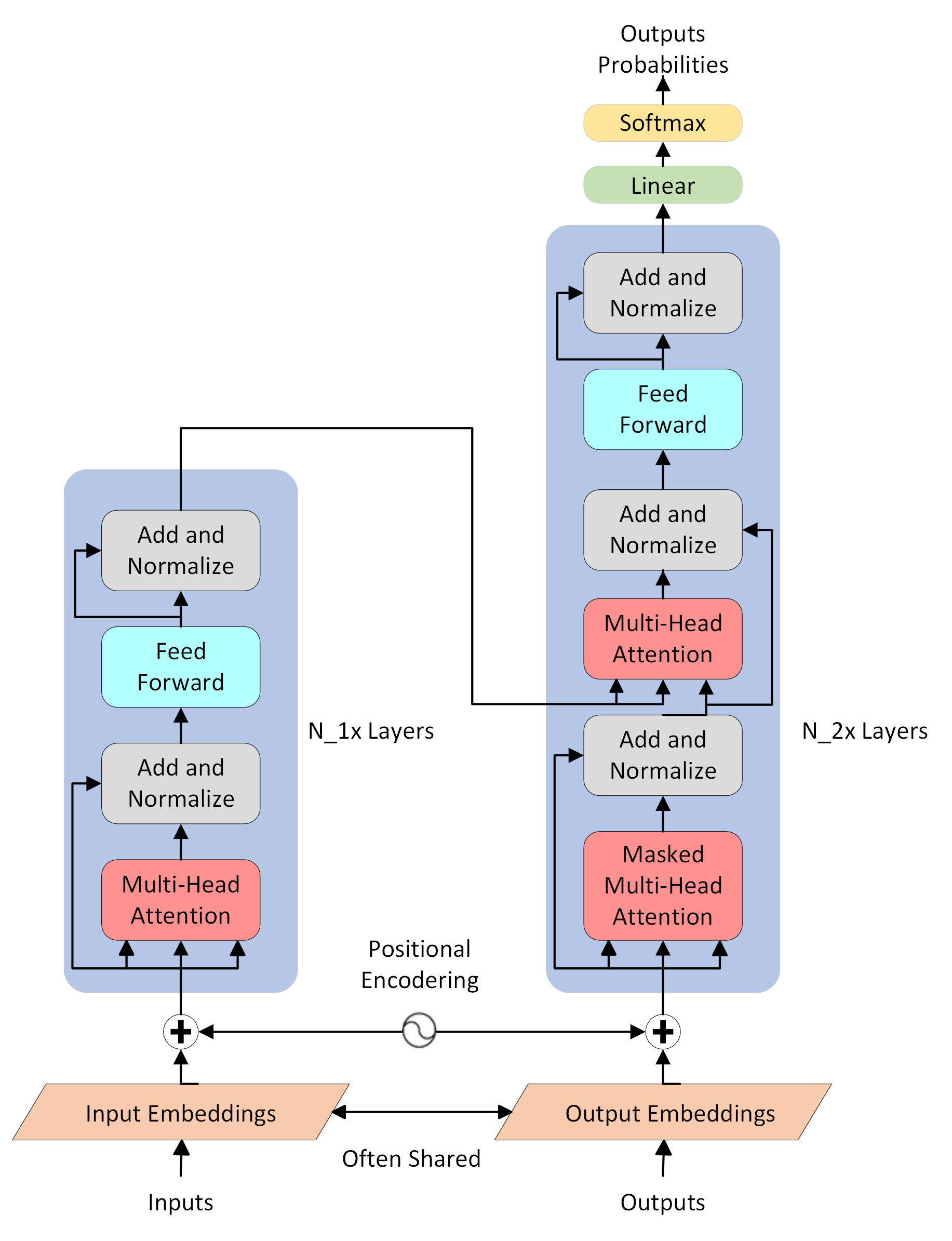}
    \caption{Depiction of classical transformer architecture as constructed by Vaswani et. al \cite{vaswani17}. Input and output embeddings are commonly shared, as shown, to reduce model size and it is typical to use similar, if not identical, positional encoding schemes to produce recurrence. While not shown here, it has become somewhat popular to rearrange the order of the normalization blocks \cite{layernorm}, placing them before the attention and feed-forward blocks to slightly speed up and stabilize training with little to no cost to end performance.}
    \label{fig:basic_transf}
\end{figure}

\begin{figure}[!htb]
    \centering
    \includegraphics[scale=0.6]{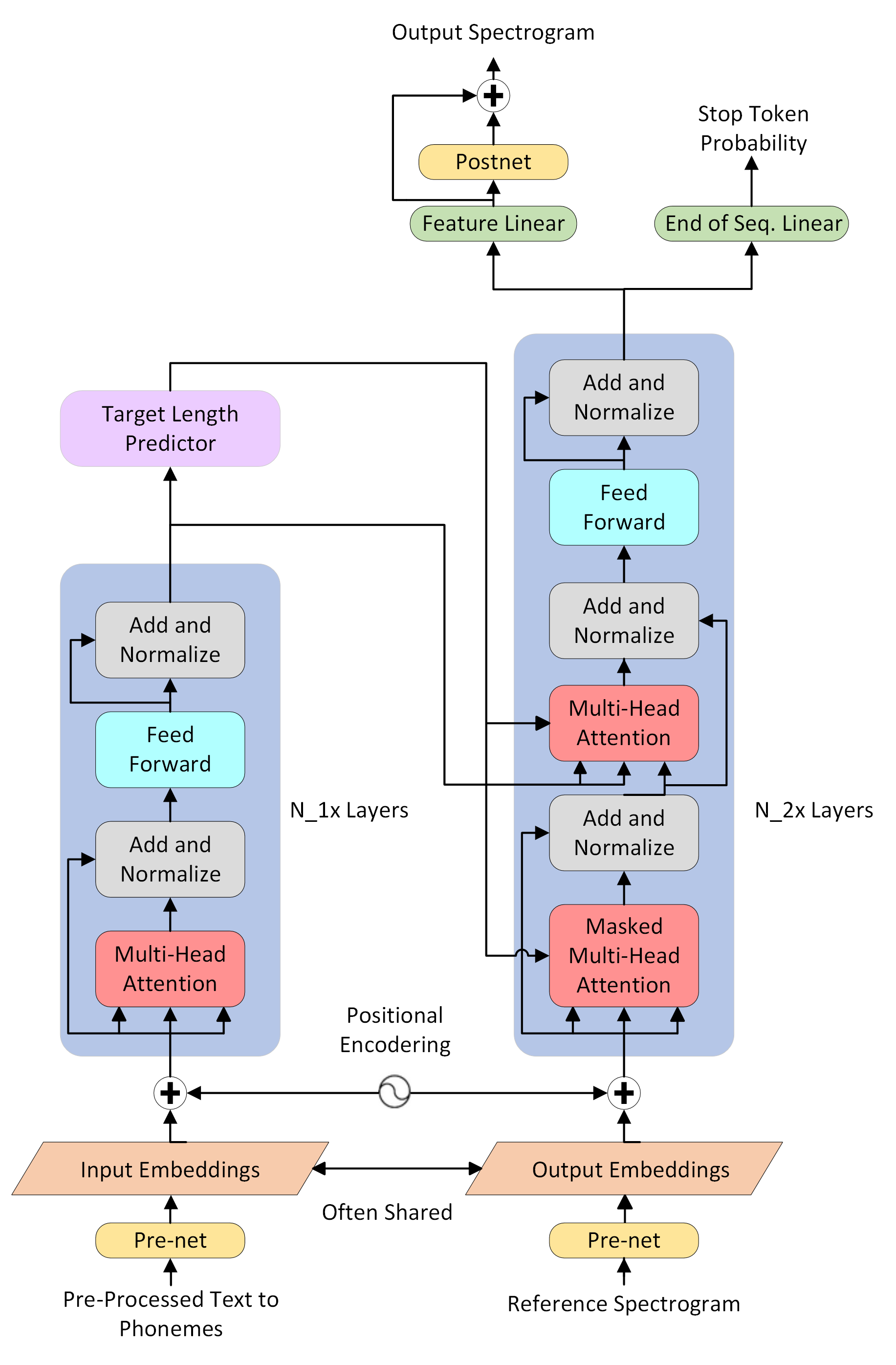}
    \caption{Depiction of TransformerTTS-based model used for the following experiments. Input and output embeddings were shared and normalization blocks were reordered to occur before other relevant blocks, despite not being shown here. The length predictor is varied in later experiments, with implementations that are inference only, implementations that are trained, and implementations that must be used during training for model robustness.}
    \label{fig:tts_mod_transf}
    \vspace{5em}
\end{figure}

Two figures are provided below illustrating Vaswani et. al's \cite{vaswani17} classical transformer and TransformerTTS \cite{ttstransformer} with the proposed target sequence length predictor augmentation found in this paper. Figure \ref{fig:basic_transf} represents a classical transformer while Figure \ref{fig:tts_mod_transf} represents a modified TransformerTTS.

\newpage

\subsection{Feature Comparisons for Synthesized Spectrograms}
\begin{figure}[!htb]
    \centering
    \includegraphics[scale=0.4]{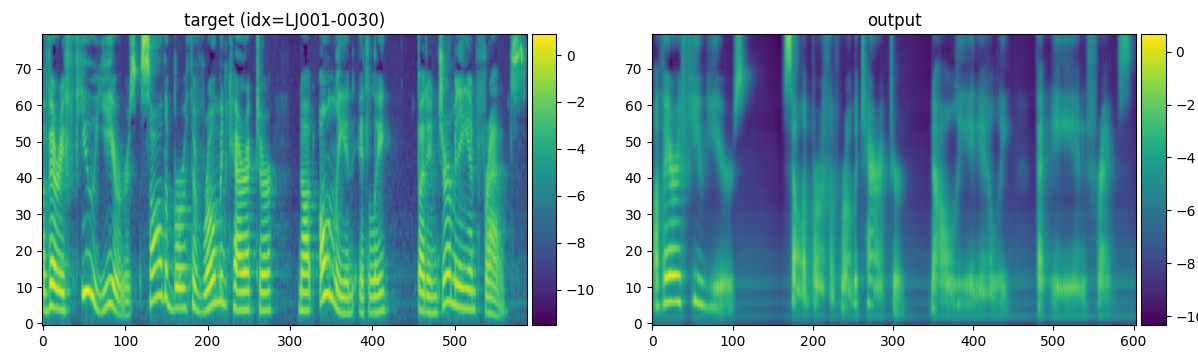}
    \caption{Spectrogram comparison between the reference spectrogram and the synthesized spectrogram from a baseline model with only softmax attention on sample LJ001-0030. This sample was randomly selected out of the test set. The left axis is the frequency in kHz, the right axis is the decibel value, and the bottom axis is the duration in frames.}
    \label{fig:30_base_crop}
\end{figure}

\begin{figure}[!htb]
    \centering
    \includegraphics[scale=0.4]{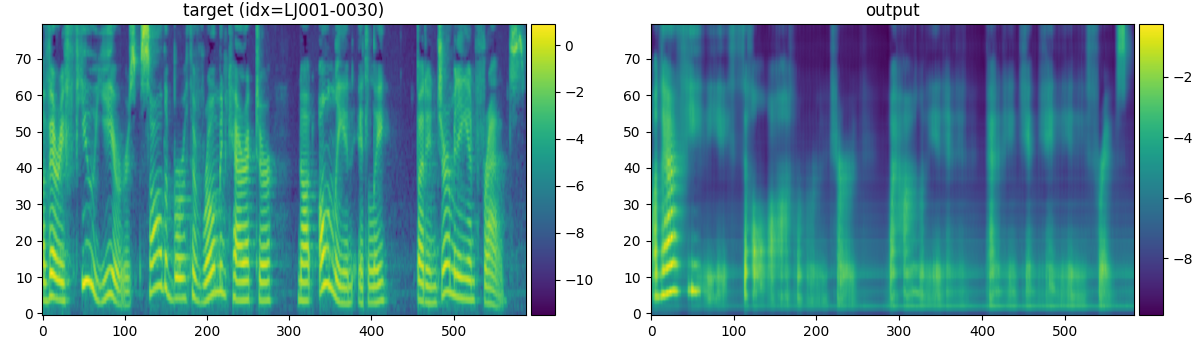}
    \caption{Spectrogram comparison between the reference spectrogram and the synthesized spectrogram from a model with cosFormer encoder to decoder cross-attention and softmax self-attention blocks on sample LJ001-0030. This sample was randomly selected out of the test set. The left axis is the frequency in kHz, the right axis is the decibel value, and the bottom axis is the duration in frames.}
    \label{fig:30_cos_crop}
\end{figure}

A handful of samples were randomly selected for comparisons between the reference spectrogram and the synthesized spectrograms. In many cases, the linearized model (only cross-attention was linearized for this comparison) was able to somewhat compete with the baseline, full softmax attention model insofar as visible features are concerned. Figure \ref{fig:30_base_crop} and Figure \ref{fig:30_cos_crop} are good examples of an instance where the linearized model seems to have performed somewhat comparably, with both capturing the starting features reasonably well. Spectrogram quality, in general, seemed to degrade for later decoding time-steps. 


\begin{figure}[!htb]
    \centering
    \includegraphics[scale=0.4]{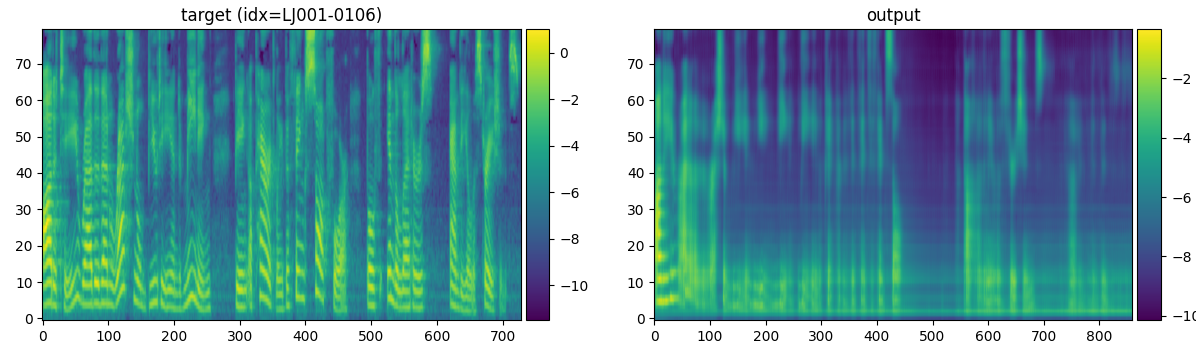}
    \caption{Spectrogram comparison between the reference spectrogram and the synthesized spectrogram from a baseline model with only softmax attention on sample LJ001-0106. This sample was randomly selected out of the test set. The left axis is the frequency in kHz, the right axis is the decibel value, and the bottom axis is the duration in frames.}
    \label{fig:106_base_crop}
\end{figure}

\begin{figure}[!htb]
    \centering
    \includegraphics[scale=0.4]{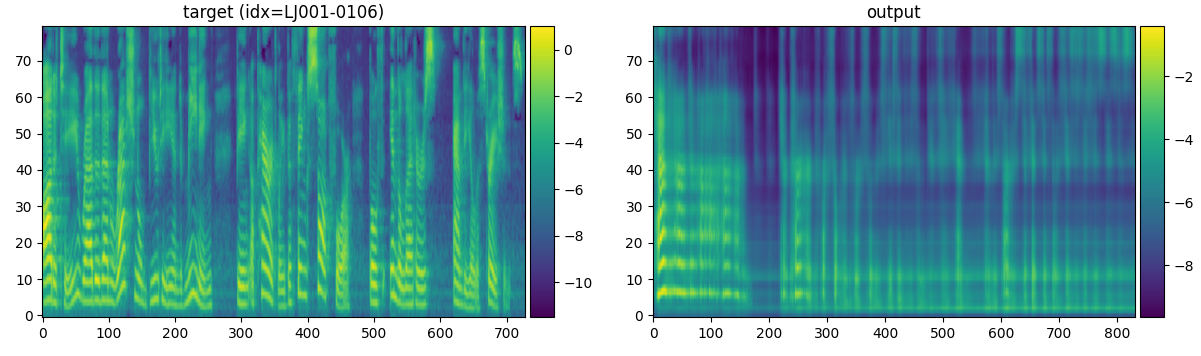}
    \caption{Spectrogram comparison between the reference spectrogram and the synthesized spectrogram from a model with cosFormer encoder to decoder cross-attention and softmax self-attention blocks on sample LJ001-0106. This sample was randomly selected out of the test set. The left axis is the frequency in kHz, the right axis is the decibel value, and the bottom axis is the duration in frames.}
    \label{fig:106_cos_crop}
    \vspace{18em}
\end{figure}


\begin{figure}[!htb]
    \centering
    \includegraphics[scale=0.4]{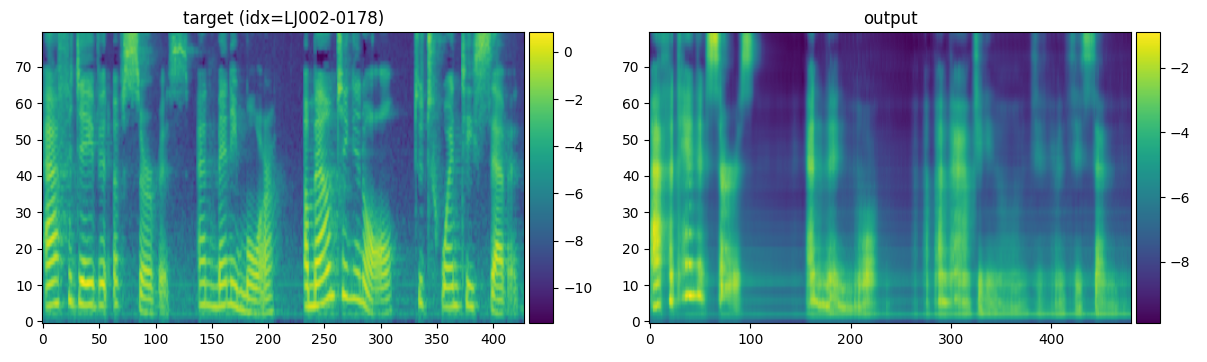}
    \caption{Spectrogram comparison between the reference spectrogram and the synthesized spectrogram from a baseline model with only softmax attention on sample LJ002-0178. This sample was randomly selected out of the test set. The left axis is the frequency in kHz, the right axis is the decibel value, and the bottom axis is the duration in frames.}
    \label{fig:178_base_crop}
\end{figure}

\begin{figure}[!htb]
    \centering
    \includegraphics[scale=0.4]{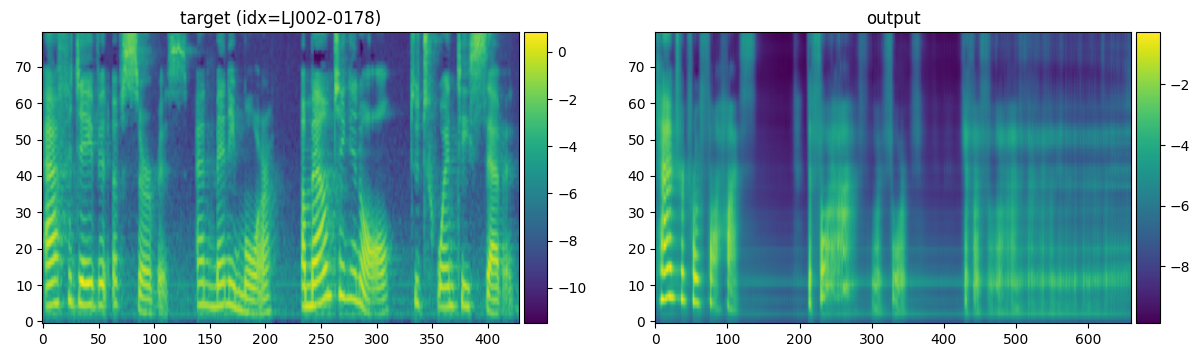}
    \caption{Spectrogram comparison between the reference spectrogram and the synthesized spectrogram from a model with cosFormer encoder to decoder cross-attention and softmax self-attention blocks on sample LJ002-0178. This sample was randomly selected out of the test set. The left axis is the frequency in kHz, the right axis is the decibel value, and the bottom axis is the duration in frames.}
    \label{fig:178_cos_crop}
\end{figure}

\newpage

\subsection{Final Decoder Layer Cross-Attention Comparisons}
A handful of samples were randomly selected for comparisons between a baseline softmax attention model and a model with its encoder to decoder cross-attention blocks replaced by a cosFormer implementation for the final layer of the decoder. To generate these scores, features were averaged across attention heads for the intermediate and normalized $QK^T$ matrix (this was generated purely for logging during linearized inference and was not used for prediction). The general shape of this layer's cross-attention does seem to be reasonably approximated by cosFormer cross-attention, but it should be noted that the diagonal structure expected of modules that cosFormer should perform well with is not truly observed here (i.e. a token from the encoder stack's output at some relative position should score strongly with tokens in the same relative position in the decoder stack for cosFormer to perform particularly well). It should be noted that cosFormer seems to be somewhat peakier for its scores, so an initial heatmap comparison looks rather strange but common features are still very much observable. Values that correspond to scores of zero are typically padding symbols.


\begin{figure}[!htb]
    \centering
    \includegraphics[scale=0.4]{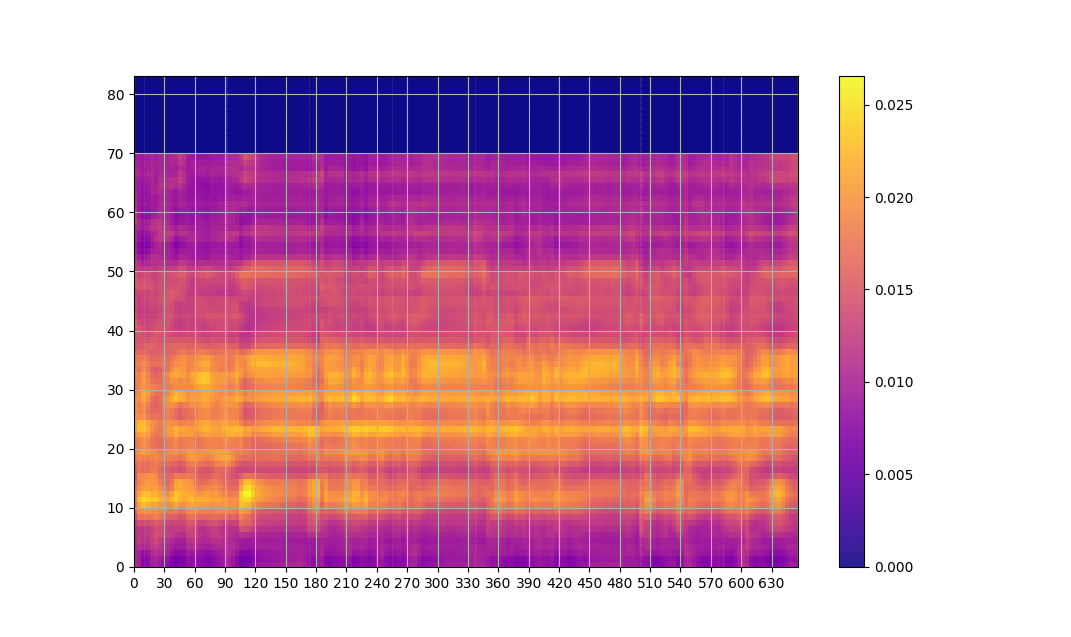}
    \caption{Softmax $QK^T$ intermediate matrix results for the cross-attention of the final layer of the decoder stack for sample LJ001-0030 in LJSpeech's test set. The left axis represents the source tokens in the form of phonemes and the bottom axis represents the predicted waveform tokens.}
    \label{fig:30_base_attn}
    \vspace{-1em}
\end{figure}

\begin{figure}[!htb]
    \vspace{-2em}
    \centering
    \includegraphics[scale=0.4]{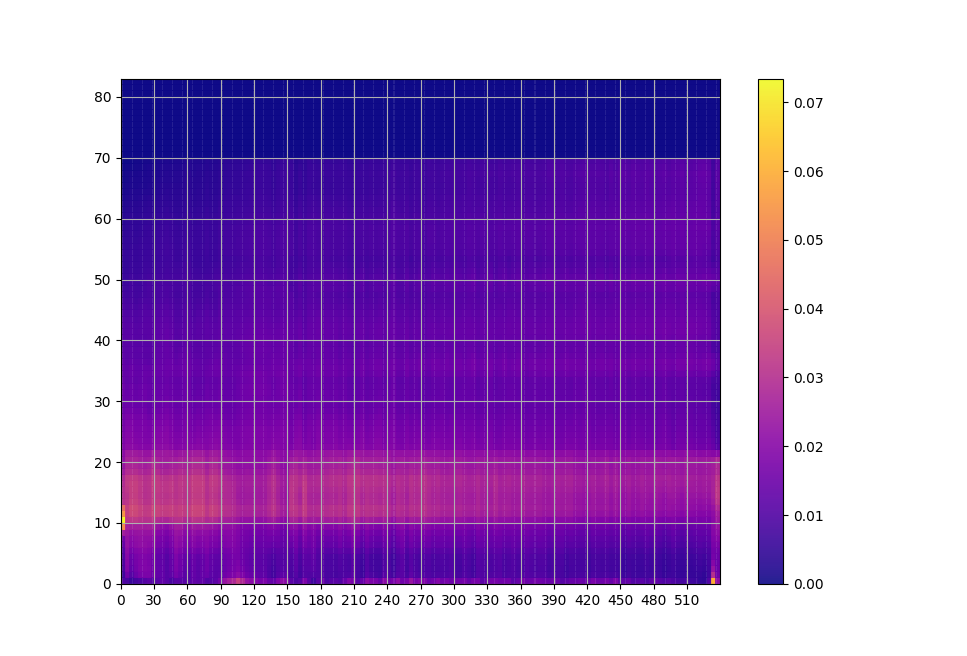}
    \caption{cosFormer $QK^T$ intermediate matrix results for the cross-attention of the final layer of the decoder stack for sample LJ001-0030 in LJSpeech's test set. The left axis represents the source tokens in the form of phonemes and the bottom axis represents the predicted waveform tokens.}
    \label{fig:30_cos_attn}
    \vspace{-1em}
\end{figure}

\begin{figure}[!htb]
    \centering
    \includegraphics[scale=0.4]{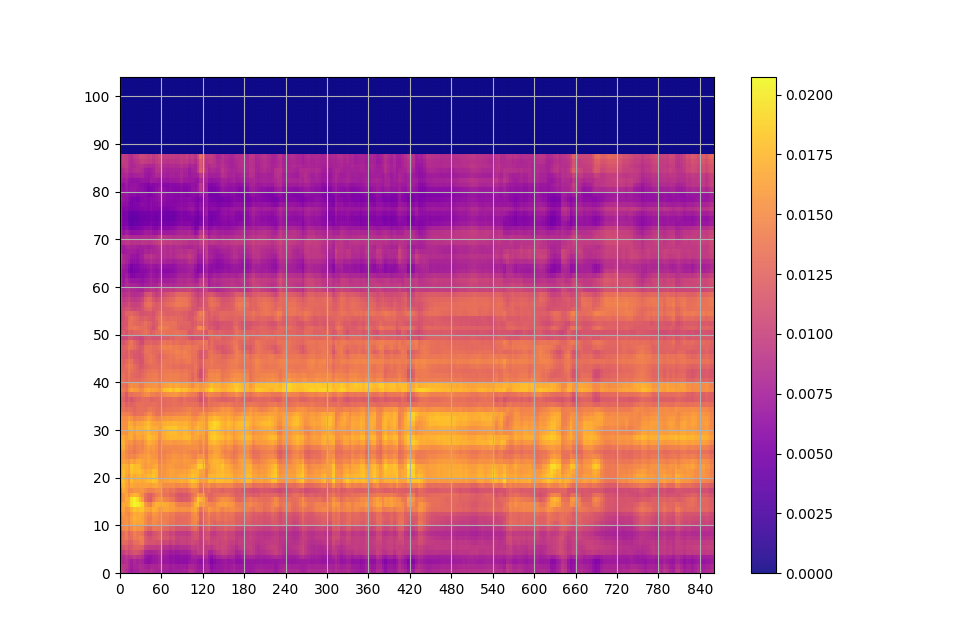}
    \caption{Softmax $QK^T$ intermediate matrix results for the cross-attention of the final layer of the decoder stack for sample LJ001-0106 in LJSpeech's test set. The left axis represents the source tokens in the form of phonemes and the bottom axis represents the predicted waveform tokens.}
    \label{fig:106_base_attn}
\end{figure}

\begin{figure}[!htb]
    \vspace{-2em}
    \centering
    \includegraphics[scale=0.35]{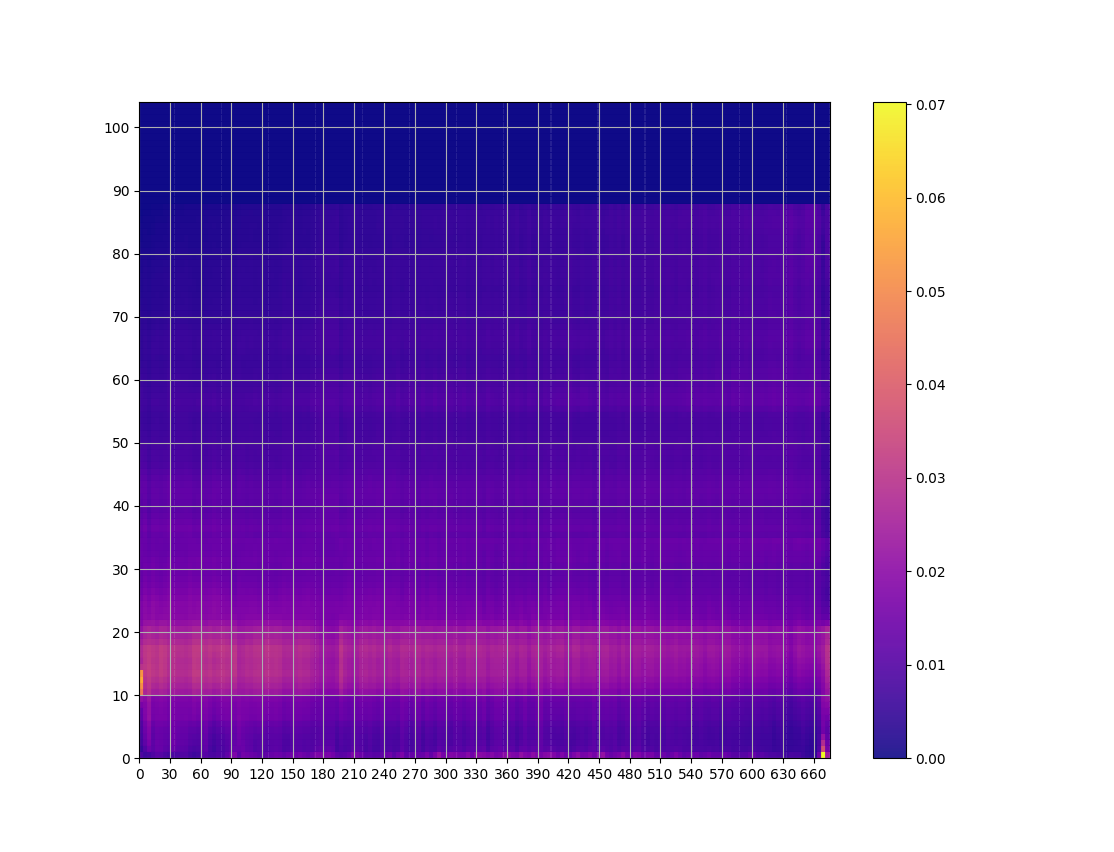}
    \caption{cosFormer $QK^T$ intermediate matrix results for the cross-attention of the final layer of the decoder stack for sample LJ001-0106 in LJSpeech's test set. The left axis represents the source tokens in the form of phonemes and the bottom axis represents the predicted waveform tokens.}
    \label{fig:106_cos_attn}
    \vspace{-2em}
\end{figure}

\begin{figure}[!htb]
    \centering
    \includegraphics[scale=0.4]{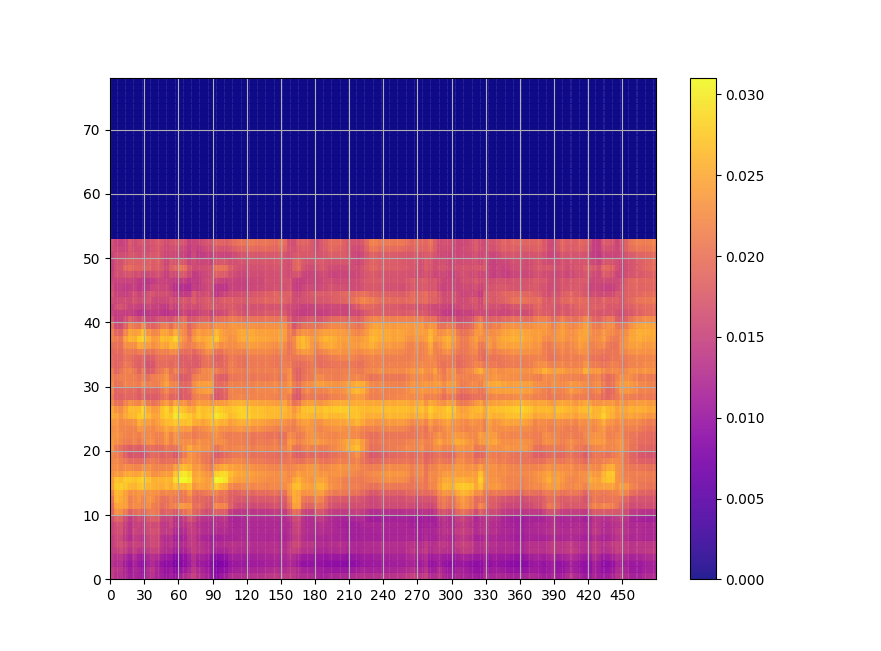}
    \caption{Softmax $QK^T$ intermediate matrix results for the cross-attention of the final layer of the decoder stack for sample LJ002-0178 in LJSpeech's test set. The left axis represents the source tokens in the form of phonemes and the bottom axis represents the predicted waveform tokens.}
    \label{fig:178_base_attn}
\end{figure}

\begin{figure}[!htb]
    \centering
    \includegraphics[scale=0.4]{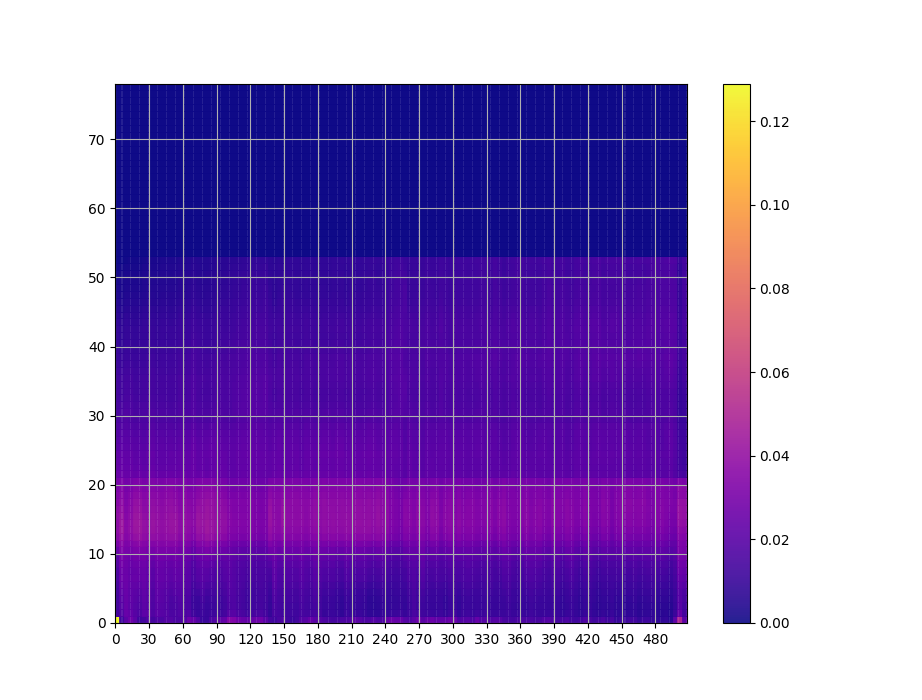}
    \caption{cosFormer $QK^T$ intermediate matrix results for the cross-attention of the final layer of the decoder stack for sample LJ002-0178 in LJSpeech's test set. The left axis represents the source tokens in the form of phonemes and the bottom axis represents the predicted waveform tokens.}
    \label{fig:178_cos_attn}
\end{figure}

\newpage

\subsection{Relevant Algorithms}
The following algorithm is somewhat relevant to this paper, but was not considered critical to include in its main body. Algorithm \ref{alg:data_reuse} provides a reference for the data-reuse implementation inspired by Katharapalous et. al's work \cite{kathrapalous20}, resulting in a run-time complexity that is linear with respect to the number of samples during inference. This particular algorithm is extremely generalizable and portions of it are cut away for the sake of efficiency given a particular linearization scheme. 

\begin{algorithm}
\caption{Generalized fully linearized attention for autoregressive decoding at inference for some arbitrary decoding time-step $m + 1$ for a single attention head. Note that keeping track of the transformed keys is optional, but can be useful under some specific circumstances.}
\label{alg:data_reuse}
\textbf{Input} \textit{Q} in $\mathbb{R}^{1 \times d}$, \textit{K} in $\mathbb{R}^{N_2 \times d}$, \textit{V} in $\mathbb{R}^{N_2 \times d}$, $M_{pr}$ in $\mathbb{R}^{d \times d}$, $S_{pr}$ in $\mathbb{R}^{1 \times d}$, \\ $K^{tr}_{pr}$ in $\mathbb{R}^{N_2 \times d}$ \\
\textbf{Output} \textit{A} in $\mathbb{R}^{1 \times d}$ \\
\textbf{Require} Flag $f$ for $K$ and $V$ updates, chunk size \textit{k} for \textit{K} and \textit{V} updates, decomposable linear similarity function defined by $S_q$ and $S_k$, linear re-weighting scheme defined by $R_q$ and $R_k$ \\
\begin{algorithmic}
    \STATE $Q' \gets S_q(Q)$
    \STATE $Q^{tr} \gets R_q(Q')$

    \IF {$f$ is True}
        \STATE $K'_{up} \gets S_k(K[-k:])$
        \STATE $K^{tr}_{up} \gets R_k(K'_{up})$
        \STATE $K^{tr} \gets concat(K^{tr}_{pr}, K^{tr}_{up})$
    \ELSE
        \STATE $K^{tr} \gets K^{tr}_{pr}$
    \ENDIF
    \STATE
    
    \IF {$f$ is True}
        \STATE $S_{up} \gets \sum K^{tr T}_{up}$
        \STATE $S \gets S_{pr} + S_{up}$
    \ELSE
        \STATE $S \gets S_{pr}$
    \ENDIF
    \STATE

    \IF {$f$ is True}
        \STATE $M_{up} \gets K^{tr T}_{up} \times V[-k:]$
        \STATE $M \gets M_{pr} + M_{up}$
    \ELSE
        \STATE $M \gets M_{pr}$
    \ENDIF
    \STATE

    \STATE $D \gets Q^{tr} \times M$
    \STATE $N \gets Q^{tr} \times S$
    \STATE $A' \gets \frac{D}{N}$
    \STATE $A \gets out(A')$  

    \STATE \RETURN $A$
\end{algorithmic}
\end{algorithm}

\newpage 

\subsection{Synthetic Dataset for Brief Latency Comparisons}
In addition to the run-time analyses for practical workloads in this paper, below is a brief analysis of the effect of embedding dimension size and sequence length on the latency of softmax attention and various linearization schemes for decoding attention blocks and a single attention head. The data used for this analysis is entirely synthetic, generated for rote calculation to demonstrate run-time profiles with no priors. CPU and GPU-based analyses are provided below, with the CPU in question being an Intel Xeon Platinum 8168 and the GPU being a single NVIDIA Tesla V100. Batching is disabled for CPU runs and enabled for some GPU runs to showcase differences in run-time profiles for practical workloads and environments.

Generally, performances on CPUs resulted in observed latency advantages for simple ReLU attention implementations and essentially no latency advantages for cosFormer. On the side of GPUs, when batching was disabled, similar behavior was observed. Only with batching enabled (e.g. something like TTS for speeches where the entire speech is available immediately is a possible example of when batching might be enabled), does cosFormer begin to demonstrate latency advantages. This is attributed to some overhead in PyTorch's under-the-hood optimizations that can render very lightweight operations cumbersome if they are not engaging in large scale, and highly parallelizable, floating point operations.

\begin{figure}[h]
    \centering
    \includegraphics[width=\textwidth]{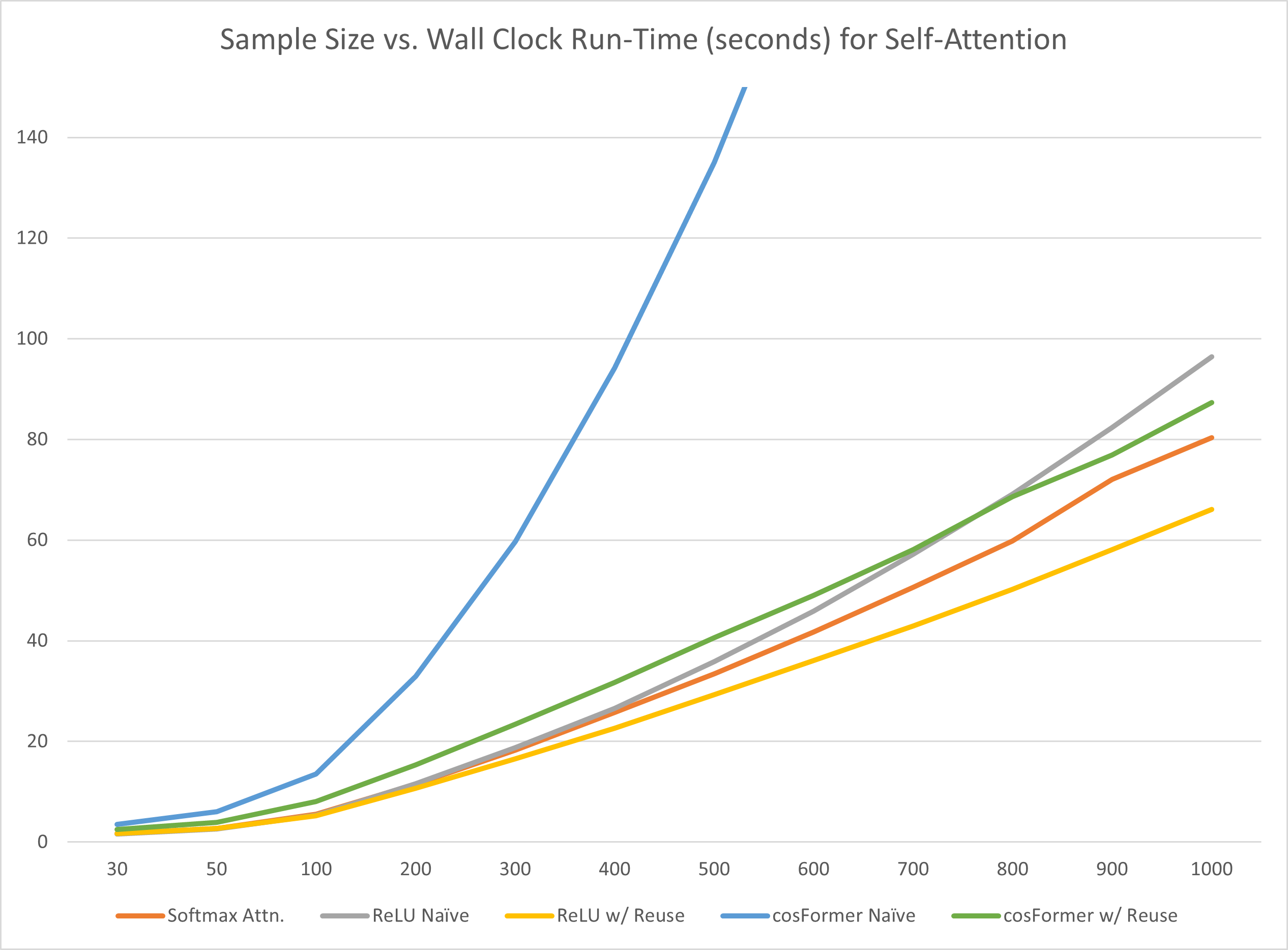}
    \caption{Comparisons of run-time profiles for decoder self-attention for varying sample lengths on a CPU. 250 samples are present in this synthetic dataset with batching disabled. Naive linearization techniques make no attempt at reusing the $K^TV$ intermediate matrix while reuse implies storing older information for later use. Notably, cosFormer does not beat out softmax implementations for any of the provided sample lengths.}
    \label{fig:cpu_self}
\end{figure}

\begin{figure}[h]
    \centering
    \includegraphics[width=\textwidth]{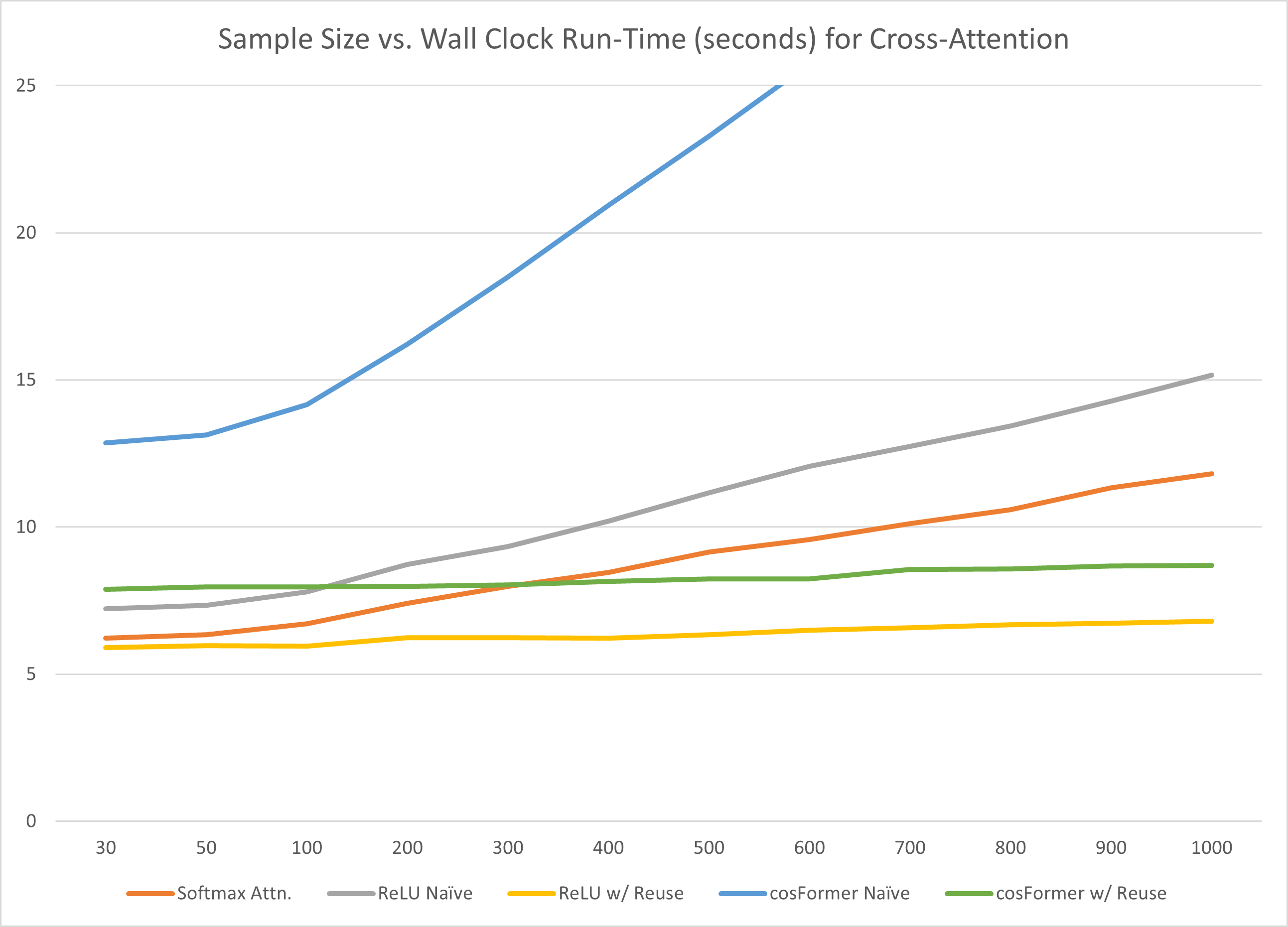}
    \caption{Comparisons of run-time profiles for decoder cross-attention for varying sample lengths on a CPU. 250 samples are present in this synthetic dataset with batching disabled. For cross-attention, the sample length variations vary the source length (key and value sizes) while the target length is set to 150 tokens. Naive linearization techniques make no attempt at reusing the $K^TV$ intermediate matrix while reuse implies storing older information for later use. Notably, cosFormer does not beat out softmax implementations at practical sample lengths for TTS.}
    \label{fig:cpu_cross}
\end{figure}

\begin{figure}[h]
    \centering
    \includegraphics[width=\textwidth]{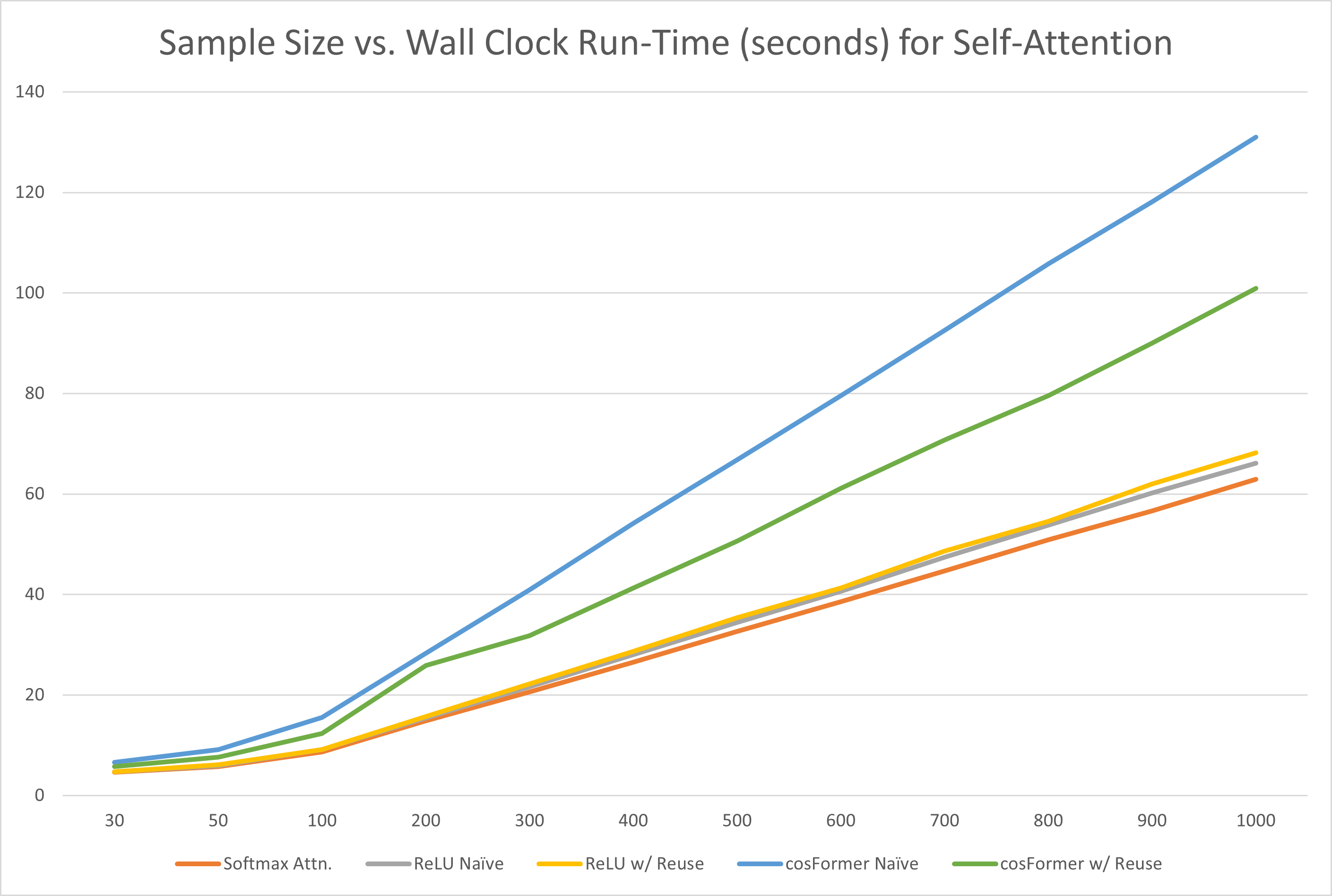}
    \caption{Comparisons of run-time profiles for decoder self-attention for varying sample lengths on a GPU. 250 samples are present in this synthetic dataset with batching disabled. Naive linearization techniques make no attempt at reusing the $K^TV$ intermediate matrix while reuse implies storing older information for later use. Notably, cosFormer does not beat out softmax implementations for any of the provided sample lengths.}
    \label{fig:gpu_self}
\end{figure}

\begin{figure}[h]
    \centering
    \includegraphics[width=\textwidth]{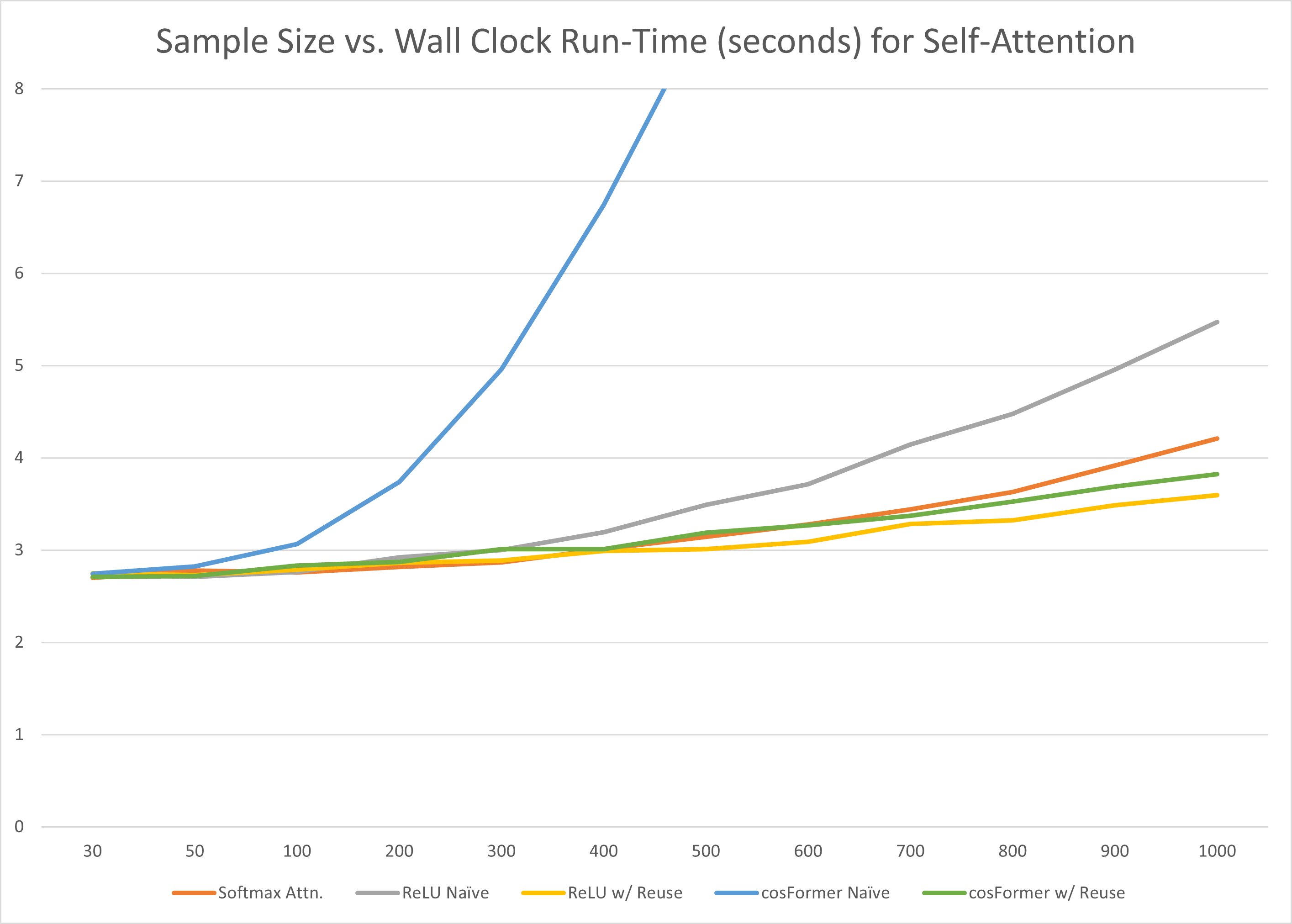}
    \caption{Comparisons of run-time profiles for decoder self-attention for varying sample lengths on a GPU. 250 samples are present in this synthetic dataset with batches of 125. Naive linearization techniques make no attempt at reusing the $K^TV$ intermediate matrix while reuse implies storing older information for later use. Notably, cosFormer does not beat out softmax implementations at practical sample lengths for TTS.}
    \label{fig:gpu_self_no_bsz}
\end{figure}

\begin{figure}[h]
    \centering
    \includegraphics[width=\textwidth]{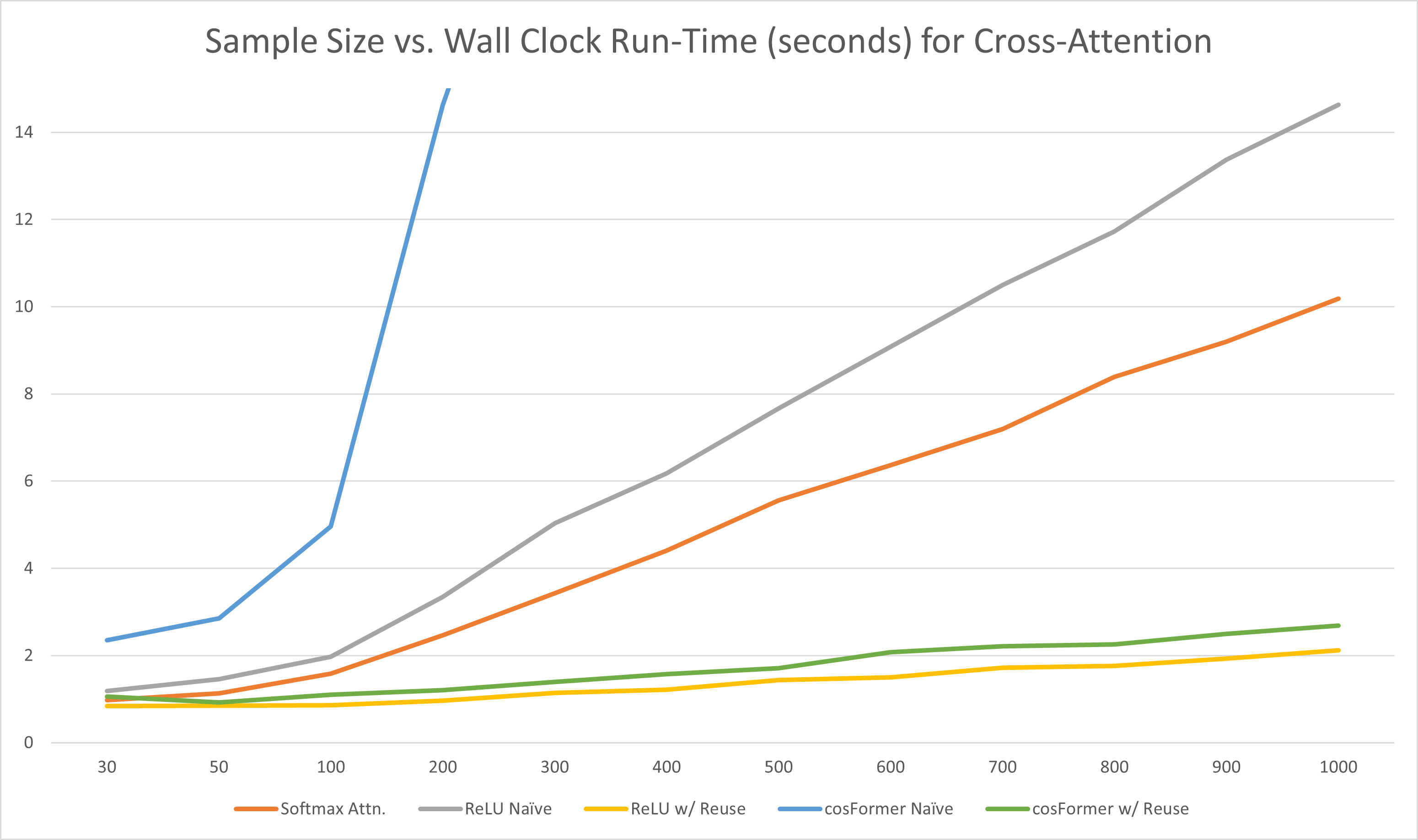}
    \caption{Comparisons of run-time profiles for decoder cross-attention for varying sample lengths on a GPU. 250 samples are present in this synthetic dataset with batches of 125. For cross-attention, the sample length variations vary the source length (key and value sizes) while the target length is set to 150 tokens. Naive linearization techniques make no attempt at reusing the $K^TV$ intermediate matrix while reuse implies storing older information for later use. Notably, cosFormer easily beats out softmax implementations at all relevant sample lengths.}
    \label{fig:gpu_self_w_bsz}
\end{figure}

\end{document}